\documentclass[lettersize,journal]{IEEEtran}
\usepackage{amsmath,amsfonts}
\usepackage{algorithm}     
\usepackage{algpseudocode} 
\usepackage{array}
\usepackage[caption=false,font=normalsize,labelfont=sf,textfont=sf]{subfig}
\usepackage{textcomp}
\usepackage{stfloats}
\usepackage{url}
\usepackage{verbatim}
\usepackage{graphicx}
\usepackage{cite}

\usepackage{makecell}
\usepackage{multirow, multicol}
\usepackage{adjustbox}
\usepackage{boldline}
\usepackage{amssymb}%
\usepackage{booktabs}
\usepackage{amsmath}
\usepackage{pifont}

\usepackage{array}
\usepackage{color}
\usepackage{hyperref}
\usepackage{xcolor} 
\usepackage{colortbl}
\usepackage{multirow}
\usepackage{makecell}
\usepackage{tabularray}
\usepackage{xcolor}

\begin{document}

\title{Distilling Image Prototypes for Guided Test-Time Adaptation}

\author{Liwen Wang, Xingbo Dong, Iman Yi Liao, Deyin Liu, \IEEEmembership{Member, IEEE}, \\Massimo Tistarelli, Lin Yuanbo Wu, \IEEEmembership{Senior Member, IEEE}, Zhe Jin, \IEEEmembership{Member, IEEE}
\thanks{This work was supported by the National Natural Science Foundation of China (Grant No. 62306003), the Open Research Fund of Guangdong Laboratory of Artificial Intelligence and Digital Economy (SZ) (Grant No. GML-KF-24-29), and the Open Foundation of Jiangxi Provincial Key Laboratory of Image Processing and Pattern Recognition (Grant No. ET202404437).  \textit{(Corresponding author: Xingbo Dong and Zhe Jin)}}
\thanks{Liwen Wang, Xingbo Dong, Deyin Liu, and Zhe Jin are with the Anhui Provincial Key Laboratory of Secure Artificial Intelligence, Anhui Provincial International Joint Research Center for Advanced Technology in Medical Imaging, School of Artificial Intelligence, Anhui University, Hefei 230093, China (e-mail: liwenwang@stu.ahu.edu.cn; xingbo.dong@ahu.edu.cn; jinzhe@ahu.edu.cn).\\ Iman Yi Liao is with the School of Computer Science, University of Nottingham (Malaysia Campus), Selangor Darul Ehsan, 43500, Malaysia  (e-mail: Iman.Liao@nottingham.edu.my).\\ Massimo Tistarelli is with the Department of Engineering, University of Sassari, Sassari, Italy  (e-mail: tista@uniss.it).\\
Lin Yuanbo Wu is with the School of Engineering, University of Warwick, CV4 7AL Coventry,
U.K. (e-mail: jolin.lwu@gmail.com).
}
}

\IEEEpubidadjcol

\markboth{Journal of \LaTeX\ Class Files,~Vol.~14, No.~8, August~2021}%
{Shell \MakeLowercase{\textit{et al.}}: A Sample Article Using IEEEtran.cls for IEEE Journals}


\IEEEpubid{0000--0000/00\$00.00~\copyright~2026 IEEE}

\maketitle

\begin{abstract}
Test-Time Adaptation (TTA) enhances the robustness of models against distribution shifts but faces two critical challenges: error accumulation from noisy pseudo-labels and catastrophic forgetting of source knowledge. 
Uncertainty-based approaches designed to mitigate error accumulation often yield overconfident or computationally expensive estimates, while strategies intended to prevent forgetting via prototype replay rely on static representations that easily become misaligned as the model adapts. To address these issues, this paper proposes a novel framework, Distilling Image Prototype for Guided Test-Time Adaptation (DIPTTA). The core of the proposed approach is the introduction of a Distill Image Prototype (DIP), a compact set of synthetic images that serves as a dynamic and regenerative anchor of source knowledge. This prototype enables a dynamic feature replay mechanism that continuously generates feature prototypes aligned with the current state of the model, thus effectively preventing catastrophic forgetting. Furthermore, the DIP anchors a source-calibrated uncertainty estimation method, which provides a less biased measure of sample reliability by leveraging stable source knowledge, thereby robustly suppressing error accumulation. Extensive experiments on multiple benchmarks demonstrate that DIPTTA significantly outperforms state-of-the-art methods, particularly under severe domain shifts. The source code is available at \href{https://github.com/LiwenWang919/DIPTTA}{https://github.com/LiwenWang919/DIPTTA}.
\end{abstract}


\section{Introduction}


\IEEEPARstart{D}{eep} learning models struggle when test data shifts from their training distribution. Test-Time Adaptation (TTA) addresses this by enabling online adaptation using unlabeled data~\cite{liang2025comprehensive,wang2025search,chen2023improved,hu2023unleashing}. However, TTA faces two major hurdles: error accumulation from noisy pseudo-labels and the catastrophic forgetting of original source knowledge.
Current strategies to combat catastrophic forgetting often involve replaying information from the source domain. A prominent line of work~\cite{dobler2023robust,chakrabarty2023santa} advocates for replaying pre-computed feature prototypes, which are class-wise average feature vectors derived from the source data. This line of research also includes methods that maintain a source-like feature bank or use nearest-neighbor retrieval~\cite{jang2022test}. However, these approaches have a critical flaw: these prototypes are computed once using the pretrained source model and remain static throughout the adaptation process. As the model updates its parameters to the target domain, its internal feature representation evolves, causing a growing misalignment between the static prototypes and the current feature space of the model. Consequently, these outdated prototypes provide a progressively potentially misleading anchor, thereby failing to prevent forgetting effectively. This fundamental limitation motivates the first research question: \textit{How can a replay-capable knowledge anchor be established that dynamically evolves with the model to alleviate catastrophic forgetting in a continually shifting environment?}



\IEEEpubidadjcol

Conversely, mitigating error accumulation typically relies on uncertainty estimation to identify and down-weight unreliable predictions. However, existing methods face a significant dilemma. Entropy-based techniques~\cite{wang2020tent, niu2023towards,wang2022continual} are prone to overconfidence, especially on out-of-distribution data, as they only capture the sharpness of the output distribution, not the true epistemic uncertainty of the model~\cite{liang2025comprehensive,xiao2024beyond,wang2025search,tian2025high}. While more principled Bayesian methods~\cite{gal2016dropout,mackay1992bayesian} can capture this uncertainty, they are frequently computationally prohibitive for online TTA. Recent efforts to improve their efficiency, such as through last-layer variational inference~\cite{kristiadi2020being} or simplified Laplace approximations~\cite{daxberger2021laplace}, still face challenges in non-stationary TTA settings. More critically, both types of estimates are inherently biased in the TTA context because they are computed solely based on the reaction of the model to the target data, which is itself non-stationary and potentially noisy. There is no stable reference point from the source domain to ground these estimates. This leads to the second research question: \textit{How can a stable source-domain anchor be leveraged to achieve unbiased uncertainty estimation, thereby reliably suppressing error accumulation?}

It is identified that the root cause of both limitations is the lack of a {dynamic, reliable, and efficient source of knowledge} during adaptation. To bridge this gap, this paper introduces a concept dubbed {Distilling Image Prototype (DIP)}. A DIP is not a static feature vector but a compact set of {synthetic images} distilled from the entire source dataset. This distillation process refines synthetic images so that a model trained on them closely resembles the performance of a model trained on the complete source data~\cite{zhao2021dataset}. The key advantage of the DIP is its {regenerability}: it exists in the image space, allowing it to be continuously processed through the {updated} feature extractor during adaptation. This enables the on-demand generation of feature prototypes that are inherently aligned with the current feature space of the model, thereby solving the misalignment issue of static prototypes.


Guided by this concept, the \textbf{D}istilling \textbf{I}mage \textbf{P}rototype for Guided \textbf{T}est-\textbf{T}ime \textbf{A}daptation ({DIPTTA}) framework is proposed. The DIP acts as a unified anchor, enabling two key innovations. First, to establish a replay-capable knowledge anchor that dynamically evolves with the model to prevent catastrophic forgetting in a continually shifting environment, the feature prototypes are regenerated for each incoming batch of target data by forwarding the DIP through the current student model. These dynamic prototypes are then used in a contrastive loss to pull the features of pseudo-labeled target samples towards their corresponding class anchors, ensuring that the model retains source knowledge without feature space distortion. Second, to leverage a stable source-domain anchor to achieve unbiased uncertainty estimation, thereby reliably suppressing error accumulation, the predictive uncertainty is computed by observing the response of the model (specifically, the classifier head) to the feature prototypes generated from the DIP. This measures the consistency between the target sample and the stable source knowledge, rather than the self-confidence of the model regarding the target sample alone. The resulting uncertainty measure is inherently unbiased and serves as a robust weight to modulate the contribution of each sample to the adaptation loss, effectively curating the learning signal. The key contributions can be summarized as follows:

\begin{itemize}
    \item We introduce the novel concept of a {Distilling Image Prototype (DIP)} for test-time adaptation. A DIP is a compact set of synthetic images that serve as a dynamic and regenerative knowledge anchor, overcoming the rigidity of static source representations and providing a principled solution to knowledge preservation in non-stationary environments.

    \item We develop a \textbf{D}ynamic \textbf{F}eature \textbf{P}rototype \textbf{R}eplay (DFPR) mechanism based on the DIP. This mechanism continuously regenerates feature prototypes aligned with the model's current state, effectively addressing catastrophic forgetting in continual test-time adaptation without access to the data of the source domain.

    \item We propose a \textbf{S}ource-\textbf{C}alibrated \textbf{U}ncertainty \textbf{E}stimation (SCUE) method that leverages the DIP to debias uncertainty quantification. By anchoring the uncertainty measure in the stable source knowledge, our method provides a more reliable signal for weighting target samples, leading to robust adaptation against error accumulation.

    \item Extensive experiments on multiple benchmarks demonstrate that our proposed DIP framework achieves state-of-the-art performance. The consistent improvements across various scenarios, especially under strong domain shifts, validate the effectiveness and robustness of our approach.
\end{itemize}

\section{Related Work}

\subsection{Test Time Adaptation}
Test-Time Adaptation (TTA) methods aim to adapt models via objectives applied directly to the test data. A dominant strategy has been {output-level adaptation}, which includes entropy minimization~\cite{wang2020tent,niu2023towards,yuan2023robust} to encourage confident predictions, and self-training~\cite{ma2024improved,goyal2022test,sinha2023test,jang2022test}, which uses high-confidence pseudo-labels for supervision. These methods are highly susceptible to error accumulation from noisy pseudo-labels, especially under significant domain shifts.

To enhance robustness, another line of work focuses on feature-level adaptation~\cite{iwasawa2021test,boudiaf2022parameter,wang2023feature,jung2023cafa}. This includes methods based on consistency regularization, such as the Mean Teacher paradigm~\cite{tarvainen2017mean,wang2023feature,zhang2025test,ye2025domain}, which stabilizes predictions by enforcing agreement between a teacher model and a student model under different augmentations. Furthermore, contrastive learning techniques~\cite{chen2022contrastive,ma2024discrepancy} have been adopted to learn a structured feature space with compact class clusters, improving feature discrimination in the target domain. While these methods promote feature invariance, they still fundamentally rely on the quality of pseudo-labels or the predictions of the teacher model, which can be unreliable for OOD samples~\cite{dobler2023robust}.

A critical challenge in TTA, particularly in the continual setting (CTTA)~\cite{gong2022note,sojka2023ar}, is {catastrophic forgetting}. To address this without accessing source data, recent methods have explored replaying source knowledge~\cite{niloy2024effective}. Notably, several approaches~\cite{dobler2023robust, chakrabarty2023santa} replay pre-computed, static feature prototypes of the source classes. However, a fundamental limitation persists: as the model adapts, its feature representation evolves, causing these statically computed prototypes to become progressively misaligned with the current feature space~\cite{wang2025decoupled}. This misalignment undermines their effectiveness as a stable knowledge anchor. Dataset distillation is rapidly becoming a research hotspot in machine learning, motivated primarily by the increasing scale of data and substantial model training overhead. By generating small yet potent synthetic datasets, this technology paves the way for more efficient learning and has demonstrated considerable value in numerous scenarios~\cite{geng2023survey}, including its application in continual learning~\cite{carta2022distilled,jin2025fedwsidd}. In this paper, dataset distillation is introduced into TTA to alleviate the problem of catastrophic forgetting; a replay mechanism for latent prototypes driven by dataset distillation is established, and its feasibility is verified through experiments. The proposed DIPTTA framework addresses the above limitations through a novel paradigm centered on a {Distilling Image Prototype (DIP)}. Unlike static feature prototypes, the DIP is a set of synthetic images that allows for the dynamic regeneration of feature prototypes aligned with the evolving model, effectively overcoming the misalignment issue. This dynamic replay mechanism is coupled with a source-calibrated uncertainty estimation method that leverages the DIP to debias uncertainty quantification, offering a more robust solution to both catastrophic forgetting and error accumulation.

\subsection{Uncertainty Quantification in Test Time Adaptation} 
Uncertainty estimation is crucial in TTA to identify unreliable predictions under domain shifts and to prevent error accumulation~\cite{dong2025certaintta,niu2022efficient,tan2025uncertainty}. Existing methods can be broadly categorized by their underlying principles. Simple predictive metrics, such as the prediction entropy utilized in methods like Tent~\cite{wang2020tent} and SHOT~\cite{liang2020we}, are computationally lightweight. However, these approaches primarily reflect the sharpness of the output distribution and often fail to account for the epistemic uncertainty of the model. This leads to overconfident and miscalibrated predictions on out-of-distribution data. {Principled Bayesian methods} offer a more formal framework for capturing epistemic uncertainty. Techniques such as MC Dropout~\cite{gal2016dropout}, model ensembles~\cite{rahaman2021uncertainty, abe2022deep}, and Bayesian neural networks~\cite{denker1990transforming,mackay1992bayesian} model distributions over parameters or predictions. While these methods can provide better-calibrated uncertainty estimates, their computational demands often hinder their application in online TTA scenarios. Other lightweight strategies include using the maximum softmax probability~\cite{hendrycks2016baseline} or energy scores~\cite{liu2020energy} as uncertainty proxies.

A more fundamental limitation common to both categories is that they estimate uncertainty based solely on the target domain and the adapting model. The model itself is changing, and the target data is non-stationary. This lack of a stable reference point means that the uncertainty estimates can be inherently biased by the shifting target distribution, lacking calibration to the source knowledge~\cite{trimmer2011decision,tan2025source}. This limitation is addressed by introducing a {source-calibrated uncertainty} framework. Instead of relying exclusively on the output of the adapting model for a target sample, the Distilling Image Prototype (DIP) is leveraged as a stable anchor representing source knowledge. This enables the estimation of the consistency of a target sample with the source domain, resulting in a less biased measure of reliability. By grounding the uncertainty quantification in the stable source representation, the proposed method provides a more robust mechanism for identifying unreliable predictions during adaptation.

\begin{figure*}[!t]
 \centering
 \includegraphics[width=0.9999\linewidth]{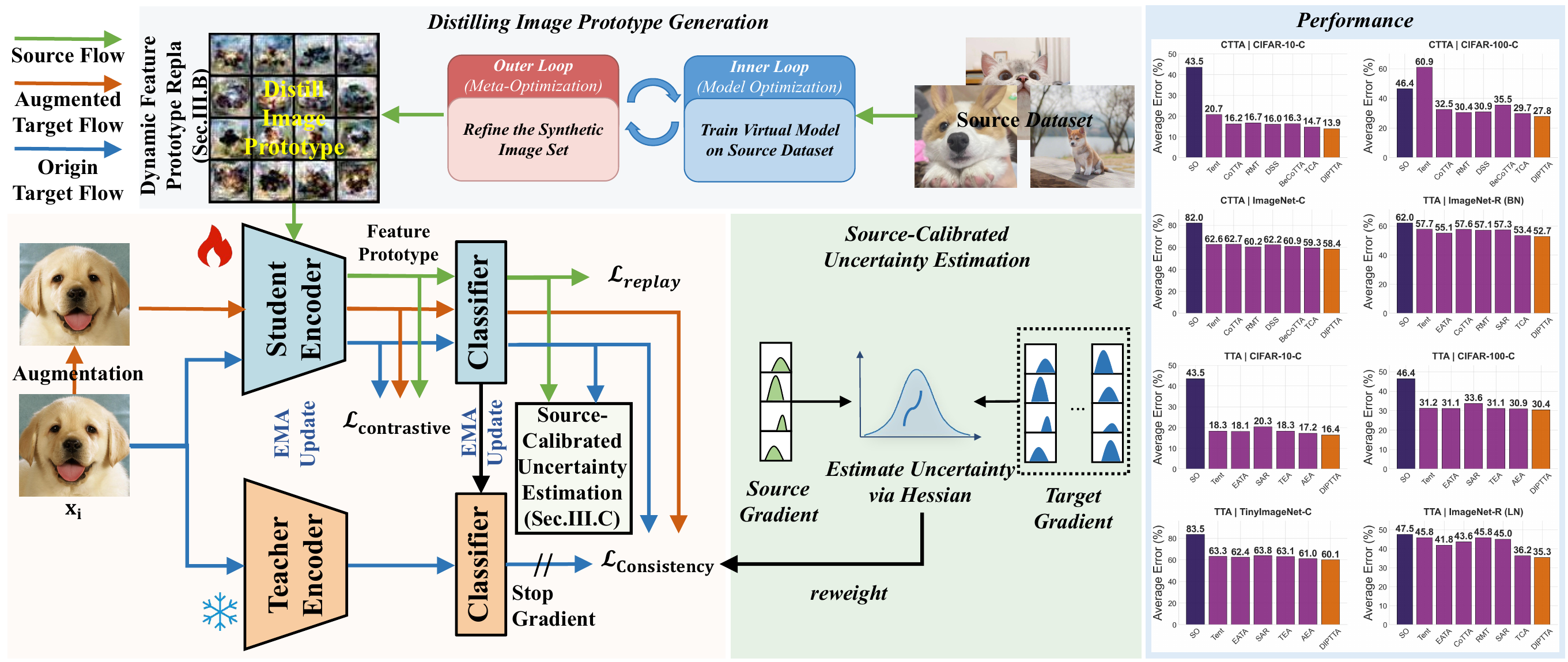}
  \vspace{-10pt}
 \caption{Overall framework of our proposed DIPTTA. The top part visualizes the Distilling Image Prototype (DIP) module, which generates high-fidelity synthetic prototypes via a two-loop distillation strategy. The bottom part depicts the core online adaptation pipeline, where a student model is continuously updated with the guidance of a teacher model. Specifically, a source-calibrated uncertainty estimation and reweighting mechanism is embedded to dynamically calibrate the gradient contribution from target data, effectively mitigating distribution shift issues in open-world adaptation scenarios.}
 \vspace{-15pt}
 \label{fig:overview}
\end{figure*}

\section{Method}
\label{sec:method}

In this section, we detail our method, Distill Image Prototype-guided Test-Time Adaptation (DIPTTA). 

\subsection{Notation and Model Architecture}
\label{3.1}

Let $\mathcal{D}_\mathcal{S} = \{(x_i^\mathcal{S}, y_i^\mathcal{S})\}$ be the source dataset, with samples from a source domain distribution $P_\mathcal{S}(x,y)$. Similarly, $\mathcal{D}_\mathcal{T} = \{(x_i^\mathcal{T})\}$ is the unlabeled target dataset from a target domain distribution $P_\mathcal{T}(x)$. The fundamental challenge in this setting is that the source distribution is different from the target distribution, i.e., $P_\mathcal{S}(x,y) \neq P_\mathcal{T}(x)P(y|x)$, although they share a common label space $\mathcal{Y}$. Given a well-trained model $M_\mathcal{S}$ pretrained on the full source data $\mathcal{D}_\mathcal{S}$, the goal is to adapt $M_\mathcal{S}$ to the target distribution under the test-time adaptation setting, which adapts the model using sequential unlabeled target batches without access to source data.

\subsubsection{Model Architecture}
To achieve this, a mean teacher framework is employed. At the start ($k=0$), both the student model ($M_{st}=h_{st}(f_{st}(\cdot))$) and the teacher model ($M_{te}=h_{te}(f_{te}(\cdot))$) are initialized with the parameters pretrained in the source domain, such that $M_{st,k=0} = M_{te,k=0} = M_\mathcal{S}$. For each incoming batch $B_k$, the parameters of the student model $\theta_{st}$ are optimized. Subsequently, the parameters of the teacher model $\theta_{te}$ are updated via an Exponential Moving Average (EMA) of the parameters of the student model:
\begin{equation}
M_{te} \leftarrow \alpha M_{te} + (1 - \alpha) M_{st}
\label{eq:ema_update}
\end{equation}
This framework enables the model to dynamically adapt to non-stationary domains by utilizing a stream of unlabeled data to track distributional changes, while concurrently receiving stable supervisory signals from the teacher component.

\subsection{Dynamic Feature Prototype Replay}
\label{3.2}


To overcome the inherent limitations of static feature prototypes, the knowledge anchor is shifted from the feature space to the image space. Unlike existing methods that rely on frozen feature vectors, the DIP framework optimizes a set of synthetic images $\mathcal{P}_{V}$ to encapsulate the essential geometry of the source distribution. This design choice is grounded in the principle of \textit{representation invariance}: while the feature extractor $f_{st}$ undergoes significant drift during target adaptation, the semantic content within the image space remains a constant reference point. By regenerating features through the latest model state (Eq.~\ref{eq:prototype}), it is ensured that the source knowledge anchor is consistently aligned with the current model, thus theoretically eliminating the misalignment error that plagues conventional replay mechanisms.

\subsubsection{Distilling Image Prototype Initialization and Optimization}
Before optimization, a small set of synthetic images $\mathcal{P}_\mathcal{V} = \{(s_1, y_1), \dots, (s_n, y_n)\}$ with $n \ll |\mathcal{D}_\mathcal{S}|$ is randomly initialized and subsequently optimized for each class. This procedure is a bi-level optimization process. 

In the Simulated Training (Inner Loop), a virtual model $M_\mathcal{V}$ (a copy of $M_\mathcal{S}$) is updated using a mini-batch from $\mathcal{P}_\mathcal{V}$ to minimize a simulated training loss, typically on the classifier head or a few top layers, thereby mimicking lightweight adaptation. The inner loop optimization can be formulated as:
\begin{equation}
\theta^{(t+1)} = \theta^{(t)} - \alpha \nabla_\theta \mathcal{L}_{\text{inner}}(B_\nu;\theta^{(t)}),
\end{equation}
where $\theta^{(t)}$ represents the model parameters at inner loop step $t$, $\alpha$ is the learning rate of the inner loop, $B_\nu$ is a mini-batch sampled from the synthetic dataset $\mathcal{P}_\mathcal{V}$, and $\mathcal{L}_{\text{inner}}$ is typically the cross-entropy loss defined as 
\begin{equation}
\mathcal{L}_{\text{inner}} = \frac{1}{|B_\nu|} \sum_{(s_i, y_i) \in B_\nu} \text{CE}(M_\mathcal{V}(s_i), y_i).
\end{equation}
In the Meta-Optimization (Outer Loop), the synthetic images $s_i$ in $\mathcal{P}_\mathcal{V}$ are optimized based on the performance of the virtually trained $M_\mathcal{V}$ on a mini-batch from the source dataset $\mathcal{D}_\mathcal{S}$. The meta-loss $\mathcal{L}_{\text{meta}}$ drives the synthetic images to encapsulate the essential knowledge of $\mathcal{D}_\mathcal{S}$. The outer loop optimization can be formulated as:
\begin{equation}
s_i \leftarrow s_i - \beta \nabla_{s_i} \mathcal{L}_{\text{meta}}(B_\mathcal{S};\theta^*(s_i)),
\end{equation}
where $s_i$ denotes the synthetic images being optimized, $\beta$ is the outer loop learning rate, $\theta^*(s_i)$ represents the model parameters after inner loop training on $\mathcal{P}_\mathcal{V}$, $B_\mathcal{S}$ is a mini-batch sampled from the source dataset $\mathcal{D}_\mathcal{S}$, and $\mathcal{L}_{\text{meta}}$ is generally the cross-entropy loss defined as 
\begin{equation}
\mathcal{L}_{\text{meta}} = \frac{1}{|B_\mathcal{S}|} \sum_{(x_j, y_j) \in B_\mathcal{S}} \text{CE}(M_\mathcal{V}(x_j), y_j).
\end{equation}
This optimization can include matching feature distributions between $M_\mathcal{V}$ on $\mathcal{P}_\mathcal{V}$ and $M_\mathcal{S}$ on $\mathcal{D}_\mathcal{S}$, and ensuring classifier consistency. Optional privacy-friendly regularization (pixel value constraints and smoothness regularization) can be applied during optimization to avoid generating images that are highly similar to the original samples.

\subsubsection{Dynamic Feature Prototype Generation}
After optimization, the final synthetic images $\mathcal{P}_\mathcal{V}^*$ are obtained. During test-time adaptation, feature prototypes are dynamically generated by passing $\mathcal{P}_\mathcal{V}^*$ through the feature extractor $f_{st}^{(k)}$ of the \textit{current} student model at each adaptation step $k$. For each class $c$, the feature prototype $\mathcal{P}_{f}^{(k)}[c]$ is computed as:
\begin{equation}
\mathcal{P}_{f}^{(k)}[c] = \textit{Mean}(\{f_{st}^{(k)}(s) \mid (s, y) \in \mathcal{P}_\mathcal{V}^*, y=c\})
\label{eq:prototype}
\end{equation}

These dynamically generated feature prototypes $\mathcal{P}_{f}^{(k)}$ are utilized in the proposed contrastive learning strategy. Given the features $f_{st}(x_i)$ extracted by the student model for a target sample $x_i$, the similarity between $f_{st}(x_i)$ and all feature prototypes is computed. The contrastive loss for a single sample $x_i$ is defined as:
\begin{equation}
\begin{aligned}
\mathcal{L}_{\text{contrastive}}(x_i) = & -\log \frac{\exp(\textit{sim}(f_{st}(x_i), \mathcal{P}_{f}^{(k)}[\hat{y}_i]) / \tau)}{\sum_{c \in C} \exp(\textit{sim}(f_{st}(x_i), \mathcal{P}_{f}^{(k)}[c]) / \tau)},
\end{aligned}
\end{equation}
where $\hat{y}_i$ is the pseudo-label predicted by the teacher model for $x_i$, $\textit{sim}(\cdot, \cdot)$ denotes cosine similarity, and $\tau$ is a temperature parameter.

This contrastive learning approach alleviates feature uncertainty in the target domain by utilizing dynamically updated feature prototypes $\mathcal{P}_{f}^{(k)}$. The prototypes are driven by the DIP $\mathcal{P}_\mathcal{V}^*$, which serves as a compact substitute for source data. The contrastive loss guides the model by drawing the features of target samples closer to their assigned class prototypes while pushing them away from other prototypes, effectively anchoring the target feature space to robust source knowledge.

\subsection{Source-Calibrated Uncertainty Estimation}
\label{3.3}

To mitigate error accumulation from noisy pseudo-labels, a source-calibrated uncertainty estimation method is proposed. Unlike traditional approaches that rely solely on target data, the proposed method uses the DIP as a stable anchor from the source domain to provide unbiased uncertainty quantification.

The source-calibrated uncertainty stems from the modeling of uncertainty sources. Traditional methods estimate uncertainty based solely on $p(y | x_t, \theta)$, which conflates aleatoric uncertainty (from target data noise) and epistemic uncertainty (from model ignorance due to domain shift). During Test-Time Adaptation (TTA), parameters $\theta$ evolve rapidly, making epistemic uncertainty difficult to isolate.

The proposed method introduces a source anchor $z_s$ (from DIP) to disentangle this conflation by evaluating the conditional distribution $p(y | x_t, z_s, \theta)$. Within a Bayesian framework:
\begin{equation}
p(y | x_t, z_s, \theta) \propto p(y | x_t, \theta) \cdot p(z_s | y, x_t, \theta)
\label{eq:bay}
\end{equation}
Here, $p(z_s | y, x_t, \theta)$ serves as a calibration factor, measuring the plausibility of the source anchor $z_s$ given the target sample and its prediction. If a target sample has high prediction confidence $p(y | x_t, \theta)$ but its features diverge from the source anchor (low $p(z_s | y, x_t, \theta)$), the calibrated probability $p(y | x_t, z_s, \theta)$ decreases, indicating potential miscalibration. Conversely, good alignment with the source anchor increases the calibrated confidence~\cite{kumar2019verified}.

Thus, the source-calibrated uncertainty, $\mathcal{U}(x_t) = \mathbb{H}[p(y | x_t, z_s, \theta)]$, corrects the biased distribution from target data alone, providing a more reliable measure of predictive uncertainty that is robust to domain shift.

While Eq.~\ref{eq:bay} defines the ideal source-calibrated distribution, directly computing the integral over the entire parameter space is computationally intractable for online TTA. To bridge this gap, the response of the model to the DIP-based prototypes is treated as a proxy for epistemic uncertainty. Specifically, the focus is placed on the uncertainty of the classification head $\theta_{h}$, as the feature extractor $f_{st}$ is assumed to be a deterministic mapping. By applying a Laplace Approximation (LA) centered at the MAP estimate $\theta_{\text{MAP}}$, the abstract calibration problem is transformed into a practical weight estimation task (Eq.~\ref{eq:entropy}), thereby anchoring the target prediction reliability in the stability of the pre-acquired source knowledge.

A direct implementation of the aforementioned theory is difficult. Therefore, the paper adopts an approximation method based on Bayesian inference, which focuses on modeling the uncertainty of the model parameters and thereby indirectly achieves source domain calibration. The posterior distribution over the network parameters $\theta$, conditioned on the source knowledge represented by the DIP-based prototypes $\mathcal{P}_{f}$, is given by $p(\theta|\mathcal{P}_{f}) \propto p(\theta)p(\mathcal{P}_{f}|\theta)$. As computing this posterior is intractable, a Laplace Approximation (LA) is employed, which forms a Gaussian distribution around a mode of the posterior, $\theta_{\text{MAP}}$:
\begin{equation}
p(\theta| \mathcal{P}_{f}) \approx \mathcal{N}(\theta|\theta_{\text{MAP}}, H^{-1})
\end{equation}
Here, $H$ is the negative Hessian of the log-posterior evaluated at $\theta_{\text{MAP}}$. To maintain computational feasibility, this Bayesian treatment is applied only to the final classification layer of the student model, $h_{st}$, using the last-layer Laplace approximation. The feature extractor $f_{st}$ remains deterministic. For efficiency, the Hessian is approximated using Kronecker-factored Laplace Approximation (KFLA)~\cite{ritter2018scalable} where $H \approx V \otimes U$. This enables the efficient estimation of the predictive posterior distribution for a given feature representation $z$:
\begin{equation}
p(y|z, \mathcal{P}_{f}) \approx \int \textit{softmax}(h_{st}(z;\theta)) p(\theta| \mathcal{P}_{f}) d\theta
\end{equation}

To derive a source-calibrated uncertainty weight $w_i$ for each target sample $x_i$, the predictive posterior is approximated using Monte Carlo (MC) integration. Specifically, $\alpha$ parameter samples $\theta_j \sim \mathcal{N}(\theta|\theta_{\text{MAP}}, H^{-1})$ are drawn to compute the mean predictive probability $\bar{P}(x_i)$. The final weight is defined as the exponentiated negative entropy of this mean prediction:
\begin{equation}
w_i = \exp(-\textit{Entropy}(\bar{P}(x_i)))
\label{eq:entropy}
\end{equation}
This weighting mechanism down-weights target samples that exhibit high predictive uncertainty, thereby mitigating negative adaptation on out-of-distribution samples.

The uncertainty weights $w_i$ are applied to two loss components. First, the \textbf{Consistency Loss} ($\mathcal{L}_{\text{Consistency}}$) quantifies the agreement between the outputs of the student and teacher models:
\begin{equation}
\begin{aligned}
\mathcal{L}_{\text{Consistency}} = & \frac{1}{N} \sum_{i=1}^N w_i \cdot \textit{SCE}(M_{st}(x_i), M_{te}(x_i)) \\
& + \frac{1}{N} \sum_{i=1}^N w_i \cdot \textit{SCE}(M_{st}(\tilde{x}_i), M_{te}(x_i)),
\end{aligned}
\end{equation}
where $\tilde{x}_i$ is an augmented version of $x_i$, and $\textit{SCE}(p,q) = \frac{1}{2}(\textit{CE}(p,q) + \textit{CE}(q,p))$ is the symmetric cross-entropy.

Second, the \textbf{Contrastive Loss} ($\mathcal{L}_{\text{Contrastive}}$) incorporates uncertainty weighting:
\begin{equation}
\mathcal{L}_{\text{Contrastive}} = \frac{1}{N} \sum_{i=1}^N w_i \cdot \mathcal{L}_{\text{contrastive}}(x_i)
\end{equation}


\subsection{Overall Optimization Objective}

The total loss for DIPTTA comprises three components: the Consistency Loss ($\mathcal{L}_{\textit{Consistency}}$), the Contrastive Loss ($\mathcal{L}_{\textit{Contrastive}}$), and the Replay Loss ($\mathcal{L}_{\textit{Replay}}$). The Replay Loss is computed on the feature prototypes driven by the latent prototypes of the source domain to prevent catastrophic forgetting via the replay mechanism:
\begin{equation}
    \mathcal{L}_{\textit{Replay}} = \frac{1}{N} \sum_{i=1}^N\textit{CE}(h_{st}(\mathcal{P}_{f}[y]), y)
\end{equation}
The final optimization objective is to minimize the total loss ($\mathcal{L}_{\textit{total}}$), a weighted sum of these components:
\begin{equation}
    \mathcal{L}_{\textit{total}} = \lambda_{1}\mathcal{L}_{\textit{Replay}}  + \lambda_{2} \mathcal{L}_{\textit{Contrastive}} + \lambda_{3} \mathcal{L}_{\textit{Consistency}}
\end{equation}
where $\lambda_{1}$, $\lambda_{2}$, and $\lambda_{3}$ are hyperparameters controlling the contribution of each loss term. The losses are weighted using $\lambda_1 = 0.5$, $\lambda_2 = 0.25$, and $\lambda_3 = 0.15$ across all experiments (refer to Section~\ref{sec:Hyperparameter} and Figure~\ref{fig:hyper} for details on the hyperparameter sensitivity analysis).

\section{EXPERIMENTS}

\subsection{Datasets and Experiment Details}
Extensive experiments are conducted to demonstrate the effectiveness of the proposed approach. DIPTTA is evaluated on five benchmark tasks for continual test-time adaptation in image processing: CIFAR-10-C, CIFAR-100-C, ImageNet-C~\cite{krizhevsky2009learning}, ImageNet-R~\cite{hendrycks2021many}, and the CCC benchmark. These tasks are designed to assess the robustness of machine learning models to corruptions and disturbances in the input data.

\begin{table*}[!ht]
\centering
\caption{Classification error rate (\%) on CIFAR10-C, CIFAR100-C, and TinyImageNet-C under TTA. Bold text indicates the best.}
\vspace{-10pt}
\begin{adjustbox}{width=0.9999\textwidth}
    \resizebox{\textwidth}{!}{
    \begin{tabular}{lcccccccccccccccc}
        \toprule
        \textbf{Method} & \textbf{Gauss.} & \textbf{Shot} & \textbf{Imp.} & \textbf{Def.} & \textbf{Glass} & \textbf{Mot.} & \textbf{Zoom} & \textbf{Snow} & \textbf{Fro.} & \textbf{Fog} & \textbf{Bri.} & \textbf{Con.} & \textbf{Ela.} & \textbf{Pix.} & \textbf{JPEG} & \textbf{Mean} \\
        \midrule
        \multicolumn{17}{c}{\textbf{CIFAR10 to CIFAR10-C}} \\
        \midrule
        Source Only & 72.3 & 65.7 & 72.9 & 46.9 & 54.3 & 34.8 & 42.0 & 25.1 & 41.3 & 26.0 & 9.3 & 46.7 & 26.6 & 58.5 & 30.3 & 43.5 \\
        Tent~\cite{wang2020tent}${}_{\textcolor{gray}{\rm ~[\text{ICLR 2021}]}}$    & 24.7 & 22.3 & 32.4 & 11.6 & 32.2 & 13.0 & 11.0 & 16.0 & 16.1 & 13.1 & 7.7 & 11.2 & 22.0 & 17.2 & 23.6 & 18.3 \\
        EATA~\cite{niu2022efficient}${}_{\textcolor{gray}{\rm ~[\text{ICML 2022}]}}$   & 24.3 & 22.3 & 32.2 & 11.3 & 31.7 & 12.9 & 10.8 & 16.0 & 16.2 & 13.4 & 7.9 & 11.3 & 21.5 & 17.0 & 23.4 & 18.1 \\
        MEMO~\cite{zhang2022memo}      & 57.2 & 51.2 & 56.3 & 32.3 & 48.7 & 27.6 & 29.6 & 20.7 & 29.3 & 20.7 & 8.6 & 30.5 & 25.1 & 52.4 & 27.4 & 34.4 \\
        SAR~\cite{niu2023towards}${}_{\textcolor{gray}{\rm ~[\text{ICLR 2023}]}}$      & 28.3 & 26.0 & 35.7 & 12.7 & 34.8 & 13.9 & 12.0 & 17.5 & 17.6 & 14.9 & 8.2 & 13.0 & 23.5 & 19.5 & 27.2 & 20.3 \\
        COME~\cite{zhang2025come}      & 24.9 & 22.1 & 29.0 & 16.9 & 33.1 & 17.9 & 15.0 & 15.6 & 12.4 & 12.8 & 7.9 & \textbf{9.4} & 21.7 & 16.0 & 19.8 & 18.3 \\
        AEA~\cite{choiadaptive}        & 25.8 & 22.6 & 31.6 & 11.8 & 30.2 & 12.5 & \textbf{10.2} & \textbf{14.5} & \textbf{14.1} & 12.4 & \textbf{7.6} & 10.8 & 19.1 & 14.6 & 20.1 & 17.2 \\
        DIPTTA (Ours) & \textbf{23.9} & \textbf{20.9} & \textbf{28.0} & \textbf{11.1} & \textbf{27.3} & \textbf{12.2} & 10.6 & 14.7 & 14.3 & \textbf{11.8} & 8.4 & 10.6 & \textbf{18.9} & \textbf{14.0} & \textbf{19.5} & \textbf{16.4} \\
        \midrule
        \multicolumn{17}{c}{\textbf{CIFAR100 to CIFAR100-C}} \\
        \midrule
        Source Only & 73.0 & 68.0 & 39.4 & 29.3 & 54.1 & 30.8 & 28.8 & 39.5 & 45.8 & 50.3 & 29.5 & 55.1 & 37.2 & 74.7 & 41.2 & 46.4 \\
        Tent~\cite{wang2020tent}${}_{\textcolor{gray}{\rm ~[\text{ICLR 2021}]}}$    & 37.3 & 34.8 & 34.5 & 25.0 & 37.4 & 27.5 & 25.1 & 30.4 & 32.0 & 33.8 & 24.1 & 28.1 & 32.9 & 28.4 & 36.9 & 31.2 \\
        EATA~\cite{niu2022efficient}${}_{\textcolor{gray}{\rm ~[\text{ICML 2022}]}}$   & 37.2 & 35.1 & 34.5 & 25.0 & \textbf{37.0} & 27.5 & 25.2 & 30.4 & 31.1 & 34.6 & 23.9 & 27.7 & 32.7 & 28.4 & 36.5 & 31.1 \\
        MEMO~\cite{zhang2022memo}      & 57.5 & 53.5 & 35.9 & 28.7 & 45.5 & 29.5 & 29.3 & 33.9 & 35.7 & 46.5 & 26.6 & 38.3 & 36.9 & 52.2 & 38.8 & 39.2 \\
        SAR~\cite{niu2023towards}${}_{\textcolor{gray}{\rm ~[\text{ICLR 2023}]}}$      & 40.5 & 37.9 & 38.7 & 26.6 & 39.9 & 28.9 & 27.0 & 33.4 & 33.2 & 38.7 & 25.5 & 29.4 & 34.2 & 31.4 & 39.4 & 33.6 \\
        COME~\cite{zhang2025come}${}_{\textcolor{gray}{\rm ~[\text{ICLR 2025}]}}$      & 37.3 & 34.8 & 34.4 & 25.0 & 37.4 & 27.6 & 25.0 & 30.4 & 31.9 & 33.7 & 24.0 & 28.1 & 32.9 & 28.3 & 36.9 & 31.1 \\
        AEA~\cite{choiadaptive}${}_{\textcolor{gray}{\rm ~[\text{ICLR 2025}]}}$        & \textbf{36.9} & 34.4 & \textbf{33.8} & \textbf{24.8} & 37.4 & 27.2 & 24.6 & \textbf{30.0} & \textbf{31.1} & 33.7 & 23.5 & 28.1 & 32.7 & \textbf{28.3} & 36.4 & 30.9 \\
        DIPTTA (Ours) & 38.8 & \textbf{32.5} & 37.5 & 25.0 & 40.9 & \textbf{24.4} & \textbf{23.7} & 30.3 & 33.5 & \textbf{32.8} & \textbf{19.1} & \textbf{21.6} & \textbf{24.7} & 31.9 & \textbf{29.2} & \textbf{30.4} \\
        \midrule
        \multicolumn{17}{c}{\textbf{ImageNet to TinyImageNet-C}} \\
        \midrule
        Source Only & 96.6 & 95.1 & 97.2 & 92.5 & 92.2 & 77.8 & 78.5 & 81.9 & 78.1 & 89.5 & 77.8 & 98.3 & 69.5 & 71.9 & 55.6 & 83.5 \\
        Tent~\cite{wang2020tent}${}_{\textcolor{gray}{\rm ~[\text{ICLR 2021}]}}$    & 66.5 & 64.3 & 72.9 & 63.2 & 75.1 & 55.5 & 54.8 & 63.6 & 61.9 & 67.6 & 57.6 & 86.2 & 56.6 & 51.5 & 52.5 & 63.3 \\
        EATA~\cite{niu2022efficient}${}_{\textcolor{gray}{\rm ~[\text{ICML 2022}]}}$   & 65.1 & 63.6 & 69.7 & 62.3 & 74.2 & 55.3 & 54.1 & 61.3 & 61.0 & 63.9 & 55.4 & 90.8 & 56.1 & 51.1 & 52.2 & 62.4 \\
        SAR~\cite{niu2023towards}${}_{\textcolor{gray}{\rm ~[\text{ICLR 2023}]}}$      & 67.2 & 65.2 & 73.6 & 63.9 & 75.9 & 55.8 & 55.2 & 64.1 & 62.6 & 68.4 & 58.0 & 86.0 & 57.1 & 51.8 & 52.9 & 63.8 \\
        COME~\cite{zhang2025come}${}_{\textcolor{gray}{\rm ~[\text{ICLR 2025}]}}$      & 66.1 & 64.6 & 72.3 & 63.2 & 75.1 & 55.8 & 54.9 & 63.1 & 61.2 & 65.9 & 56.9 & \textbf{85.3} & 56.9 & 51.8 & 52.9 & 63.1 \\
        AEA~\cite{choiadaptive}${}_{\textcolor{gray}{\rm ~[\text{ICLR 2025}]}}$        & 63.7 & \textbf{62.1} & 67.9 & \textbf{60.9} & 72.8 & 53.8 & 52.9 & 60.1 & 58.9 & 61.9 & 54.2 & 90.3 & \textbf{54.5} & 49.8 & 51.1 & 61.0 \\
        DIPTTA (Ours) & \textbf{63.4} & 62.8 & \textbf{66.8} & 65.4 & \textbf{68.8} & \textbf{53.3} & \textbf{51.5} & \textbf{58.3} & \textbf{58.2} & \textbf{58.6} & \textbf{53.9} & 90.1 & 56.2 & \textbf{47.9} & \textbf{50.1} & \textbf{60.1} \\
        \bottomrule
    \end{tabular}}
    \end{adjustbox}
  \label{tab:tta}
  \vspace{-10pt}
\end{table*}


In all experiments, the TTA setup is strictly adhered to, wherein no source data is accessed. All models are evaluated online, based on a maximum corruption severity level of five. Similar to CoTTA, standard pre-trained WideResNet~\cite{zagoruyko2016wide}, ResNeXt-29~\cite{yin2019fourier}, and ResNet-50~\cite{croce2020robustbench} are employed as the source models on a single RTX3090 (24GB) GPU for CIFAR10-C, CIFAR100-C, ImageNet-C, TinyImageNet-C, ImageNet-R, and the CCC benchmark. All results are evaluated with a corruption severity level of 5 in an online manner. For test time adaptation, the learning rate is set to 0.00025/0.001 for ResNet50/ViT-Base experiments. SGD~\cite{bottou2010large} is utilized as the optimizer, with a momentum of 0.9 and a batch size of 64.

\subsection{Comparison with SOTA methods for TTA}
Table~\ref{tab:tta} presents the comparative results of DIPTTA with state-of-the-art methods, including entropy minimization and test-time batch normalization under the setting of TTA.

\subsubsection{Comparison on CIFAR10 to CIFAR10-C}
The experimental results on the CIFAR10-C benchmark demonstrate the superiority of the proposed DIPTTA method. As shown in the table, DIPTTA achieves the lowest mean classification error of 16.4\%, outperforming the runner-up method AEA (17.2\%) by 0.8\% and significantly reducing the error compared to the Source Only baseline (43.5\%). Specifically, DIPTTA attains the best performance in 10 out of 15 corruption types, showing particular robustness in categories such as Gaussian noise, Shot noise, and Impulse noise, which clearly demonstrates the effectiveness of DIPTTA in addressing continual domain shifts and its advantage over the state-of-the-art methods.

\subsubsection{Comparison on CIFAR100 to CIFAR100-C}
On the more challenging CIFAR100-C dataset, DIPTTA continues to demonstrate robust adaptation capabilities. It achieves a state-of-the-art mean error rate of 30.4\%, surpassing competitive baselines such as AEA (30.9\%) and EATA (31.1\%). The proposed method shows significant improvements in severe corruption scenarios, particularly excelling in Contrast (21.6\%) and Elastic transform (24.7\%), where it outperforms the second-best results by substantial margins.

\subsubsection{Comparison on ImageNet to TinyImageNet-C}
For the large-scale TinyImageNet-C benchmark, DIPTTA maintains its leadership with a mean error of 60.1\%. This represents a substantial improvement over the Source Only baseline of 83.5\% and outperforms the previous method, AEA (61.0\%). The results indicate that DIPTTA scales effectively to complex datasets, achieving the lowest error rates across a diverse range of corruptions, including Glass blur, Snow, and Pixelate.

\subsection{Comparison with SOTA methods for CTTA}

The proposed DIPTTA is compared with various CTTA benchmark methods, including entropy regularization, unsupervised contrastive learning, self-training, and pseudo-label filtering. The results are shown in Table~\ref{tab:ctta} for CIFAR10-C, CIFAR100-C, ImageNet-C, and in Table~\ref{tab:ccc} for the CCC benchmark, respectively.

\subsubsection{Comparison on CIFAR10 to CIFAR10-C}
On the CIFAR10-C benchmark, directly testing the Source-only model on the target domains yields a high average error of 43.5\%. While subsequent methods like Tent, which aids in sequential adaptation, and CoTTA, which enhances pseudo-label quality, improve upon this baseline, the proposed DIPTTA method establishes a new state-of-the-art. It achieves the lowest mean error rate of 13.9\%, significantly outperforming all competing methods, including RMT (16.7\%), BeCoTTA (16.3\%), and TCA (14.7\%). This demonstrates its superior ability to adapt to continual domain shifts in this setting.

\begin{table*}[!t]
\centering
\caption{Classification error rate (\%) on CIFAR10-C, CIFAR100-C, and ImageNet-C under CTTA. Bold text indicates the best.}
\vspace{-10pt}
\begin{adjustbox}{width=0.9999\textwidth}
    \resizebox{\textwidth}{!}{
    \begin{tabular}{lcccccccccccccccc}
        \toprule
        \textbf{Method} & \textbf{Gauss.} & \textbf{Shot} & \textbf{Imp.} & \textbf{Def.} & \textbf{Glass} & \textbf{Mot.} & \textbf{Zoom} & \textbf{Snow} & \textbf{Fro.} & \textbf{Fog} & \textbf{Bri.} & \textbf{Con.} & \textbf{Ela.} & \textbf{Pix.} & \textbf{JPEG} & \textbf{Mean} \\
        \midrule
        \multicolumn{17}{c}{\textbf{CIFAR10 to CIFAR10-C}} \\
        \midrule
        Source Only & 72.3 & 65.7 & 72.9 & 46.9 & 54.3 & 34.8 & 42.0 & 25.1 & 41.3 & 26.0 & 9.3 & 46.7 & 26.6 & 58.5 & 30.3 & 43.5 \\
        Tent~\cite{wang2020tent}${}_{\textcolor{gray}{\rm ~[\text{ICLR 2021}]}}$    & 24.8 & 20.6 & 28.6 & 14.4 & 31.1 & 16.5 & 14.1 & 19.1 & 18.6 & 18.6 & 12.2 & 20.3 & 25.7 & 20.8 & 24.9 & 20.7 \\
        CoTTA~\cite{wang2022continual}${}_{\textcolor{gray}{\rm ~[\text{CVPR 2022}]}}$  & 24.3 & 21.3 & 26.6 & 11.6 & 27.6 & 12.2 & 10.3 & 14.8 & 14.1 & 12.4 & 7.6 & 10.6 & 18.3 & 13.4 & 17.3 & 16.2 \\
        RMT~\cite{dobler2023robust}${}_{\textcolor{gray}{\rm ~[\text{CVPR 2023}]}}$     & 24.0 & 20.4 & 25.6 & 12.6 & 25.4 & 14.2 & 12.2 & 15.4 & 15.1 & 14.1 & 10.3 & 13.7 & 17.1 & 13.5 & 16.0 & 16.7 \\
        DSS~\cite{wang2024continual}${}_{\textcolor{gray}{\rm ~[\text{WACV 2024}]}}$    & 24.1 & 21.3 & 25.4 & 11.7 & 26.9 & 12.2 & 10.5 & 14.5 & 14.1 & 12.5 & 7.8 & 10.8 & 18.0 & 13.1 & 17.3 & 16.0 \\
        BeCoTTA~\cite{lee2024becotta}${}_{\textcolor{gray}{\rm ~[\text{ICML 2024}]}}$   & 22.9 & {19.1} & 26.9 & \textbf{10.2} & 27.5 & 12.7 & 10.4 & 14.7 & 14.3 & 12.4 & {7.2} & {9.4} & 20.9 & 15.2 & 20.2 & 16.3 \\
        BeCoTTA~\cite{lee2024becotta}${}_{\textcolor{gray}{\rm ~[\text{CVPR 2025}]}}$    & {22.4} & 19.2 & \textbf{23.0} & 10.8 & \textbf{23.2} & {11.6} & {9.9} & {13.1} & {13.1} & {11.8} & 7.6 & 10.7 & {16.4} & {12.0} & {15.5} & {14.7} \\
        DIPTTA (Ours) & \textbf{21.9} & \textbf{18.2} & 24.7 & 10.4 & 23.8 & \textbf{11.2} & \textbf{9.5} & \textbf{12.1} & \textbf{11.8} & \textbf{10.3} & \textbf{7.1} & \textbf{8.6} & \textbf{14.7} & \textbf{10.4} & \textbf{14.2} & \textbf{13.9} \\
        \midrule
        \multicolumn{17}{c}{\textbf{CIFAR100 to CIFAR100-C}} \\
        \midrule
        Source Only & 73.0 & 68.0 & 39.4 & 29.3 & 54.1 & 30.8 & 28.8 & 39.5 & 45.8 & 50.3 & 29.5 & 55.1 & 37.2 & 74.7 & 41.2 & 46.4 \\
        Tent~\cite{wang2020tent}${}_{\textcolor{gray}{\rm ~[\text{ICLR 2021}]}}$    & {37.2} & {35.8} & 41.7 & 37.9 & 51.2 & 48.3 & 48.5 & 58.4 & 63.7 & 71.1 & 70.4 & 82.3 & 88.0 & 88.5 & 90.4 & 60.9 \\
        CoTTA~\cite{wang2022continual}${}_{\textcolor{gray}{\rm ~[\text{CVPR 2022}]}}$  & 40.1 & 37.7 & 39.7 & 26.9 & 38.0 & 27.9 & 26.4 & 32.8 & 31.8 & 40.3 & 24.7 & 26.9 & 32.5 & 28.3 & 33.5 & 32.5 \\
        RMT~\cite{dobler2023robust}${}_{\textcolor{gray}{\rm ~[\text{CVPR 2023}]}}$     & 40.5 & 36.1 & {36.3} & 27.7 & {33.9} & 28.5 & 26.4 & {29.0} & 29.0 & 32.5 & 25.1 & 27.4 & 28.2 & {26.3} & {29.3} & 30.4 \\
        DSS~\cite{wang2024continual}${}_{\textcolor{gray}{\rm ~[\text{WACV 2024}]}}$    & 39.7 & 36.0 & 37.2 & 26.3 & 35.6 & 27.5 & 25.2 & 31.4 & 30.0 & 37.8 & 24.2 & 26.0 & 30.0 & 26.3 & 31.3 & 30.9 \\
        BeCoTTA~\cite{lee2024becotta}${}_{\textcolor{gray}{\rm ~[\text{ICML 2024}]}}$   & 42.1 & 38.0 & 42.2 & 30.2 & 42.9 & 31.7 & 29.8 & 35.1 & 33.9 & 38.5 & 27.9 & 32.0 & 36.7 & 31.6 & 39.9 & 35.5 \\
        BeCoTTA~\cite{lee2024becotta}${}_{\textcolor{gray}{\rm ~[\text{CVPR 2025}]}}$    & 38.5 & 36.0 & 36.6 & {25.8} & 34.6 & {27.2} & {25.1} & 30.5 & {27.0} & {30.1} & {24.1} & {25.7} & {27.3} & 26.6 & 30.3 & {29.7} \\
        DIPTTA (Ours) & \textbf{37.4} & \textbf{34.0} & \textbf{34.6} & \textbf{25.1} & \textbf{32.4} & \textbf{25.7} & \textbf{23.5} & \textbf{26.3} & \textbf{25.9} & \textbf{29.1} & \textbf{22.4} & \textbf{23.5} & \textbf{25.7} & \textbf{23.2} & \textbf{27.6} & \textbf{27.8} \\
        \midrule
        \multicolumn{17}{c}{\textbf{ImageNet to ImageNet-C}} \\
        \midrule
        Source Only & 97.8 & 97.1 & 98.2 & 81.7 & 89.8 & 85.2 & 78.0 & 83.5 & 77.1 & 75.9 & 41.3 & 94.5 & 82.5 & 79.3 & 68.6 & 82.0 \\
        Tent~\cite{wang2020tent}${}_{\textcolor{gray}{\rm ~[\text{ICLR 2021}]}}$    & 81.6 & 74.6 & 72.7 & 77.6 & 73.8 & 65.5 & {55.3} & 61.6 & 63.0 & 51.7 & 38.2 & 72.1 & 50.8 & 47.4 & 53.3 & 62.6 \\
        CoTTA~\cite{wang2022continual}${}_{\textcolor{gray}{\rm ~[\text{CVPR 2022}]}}$  & 84.7 & 82.1 & 80.6 & 81.3 & 79.0 & 68.6 & 57.5 & 60.3 & 60.5 & 48.3 & 36.6 & 66.1 & 47.3 & 41.2 & 46.0 & 62.7 \\
        RMT~\cite{dobler2023robust}${}_{\textcolor{gray}{\rm ~[\text{CVPR 2023}]}}$     & 80.2 & 76.4 & 74.5 & 77.1 & 74.4 & 66.2 & 57.6 & 57.0 & {59.1} & 48.0 & 39.1 & 60.6 & 47.3 & 42.5 & {43.4} & 60.2 \\
        DSS~\cite{wang2024continual}${}_{\textcolor{gray}{\rm ~[\text{WACV 2024}]}}$    & 82.3 & 78.4 & 76.7 & 81.9 & 77.8 & 66.9 & 60.9 & \textbf{50.8} & 60.9 & {47.7} & \textbf{35.4} & 69.0 & 47.5 & \textbf{40.9} & 46.2 & 62.2 \\
        BeCoTTA~\cite{lee2024becotta}${}_{\textcolor{gray}{\rm ~[\text{ICML 2024}]}}$   & 84.1 & 74.3 & \textbf{72.2} & 77.4 & \textbf{71.9} & 63.4 & \textbf{55.1} & 57.2 & 61.2 & 50.7 & 36.4 & 66.1 & 49.2 & 45.6 & 48.4 & 60.9 \\
        BeCoTTA~\cite{lee2024becotta}${}_{\textcolor{gray}{\rm ~[\text{CVPR 2025}]}}$    & {78.3} & {71.8} & 73.5 & {74.4} & 73.5 & {63.3} & 56.5 & {56.9} & 59.4 & {48.1} & 39.6 & {59.6} & {47.2} & 42.9 & {44.7} & {59.3} \\
        DIPTTA (Ours) & \textbf{76.9} & \textbf{70.3} & {72.5} & \textbf{73.8} & 72.7 & \textbf{62.2} & 56.0 & {56.1} & \textbf{58.3} & \textbf{47.4} & 38.8 & \textbf{58.4} & \textbf{46.5} & 42.0 & \textbf{43.9} & \textbf{58.4} \\
        \bottomrule
    \end{tabular}}
    \end{adjustbox}
  \label{tab:ctta}
  \vspace{-10pt}
\end{table*}

\begin{table}[h!]
\centering
\caption{Classification accuracy (\%) on CCC-benchmark, which is a long-sequence task. Bold text indicates the best.}
    \vspace{-10pt}
    \begin{adjustbox}{width=0.4599\textwidth}
    \begin{tabular}{lcccc}
        \toprule
        \textbf{Method}      & \textbf{CCC-Easy}      & \textbf{CCC-Medium}    & \textbf{CCC-Hard}      & \textbf{Average} \\
        \midrule
        CoTTA~\cite{wang2022continual}${}_{\textcolor{gray}{\rm ~[\text{CVPR 2022}]}}$ & $14.9\pm0.88$          & $7.7\pm0.43$           & $1.1\pm0.16$           & 7.9              \\
        ETA~\cite{niu2022efficient}${}_{\textcolor{gray}{\rm ~[\text{ICML 2022}]}}$    & $41.4\pm0.95$          & $1.1\pm0.43$           & $0.2\pm0.05$           & 14.2             \\
        EATA~\cite{niu2022efficient}${}_{\textcolor{gray}{\rm ~[\text{ICML 2022}]}}$   & $48.2\pm0.60$          & $35.4\pm1.02$          & $8.7\pm0.80$           & 30.8             \\
        RDumb~\cite{press2023rdumb}${}_{\textcolor{gray}{\rm ~[\text{NeurIPS 2023}]}}$    & $49.3\pm0.88$          & $38.9\pm1.40$          & $9.6\pm1.60$           & 32.6             \\
        BeCoTTA~\cite{lee2024becotta}${}_{\textcolor{gray}{\rm ~[\text{CVPR 2025}]}}$    & $49.1\pm0.35$          & $39.5\pm0.53$          & $\mathbf{10.1\pm0.22}$ & 32.9             \\
        DIPTTA (Ours)         & $\mathbf{51.2\pm0.70}$ & $\mathbf{40.7\pm0.72}$ & $9.4\pm0.57$           & \textbf{33.8}    \\
        \bottomrule
    \end{tabular}
    \end{adjustbox}
    \label{tab:ccc}
    \vspace{-10pt}
\end{table}

\begin{table*}[h!]
    \centering
    \caption{Classification error rate (\%) on ImageNet-R. Results are evaluated in the single-domain adaptation scenario. Bold text indicates the best. We use * to denote episodic adaptation and use $'$ to denote the re-implemented.}
    \vspace{-10pt}
    \label{tab:imagenet-r}
    \resizebox{\textwidth}{!}{
    \begin{tabular}{lccccccccccccc}
        \toprule
        \textbf{Model} & Source Only & Tent & EATA & MEMO* & CoTTA & RMT & TEA  & SAR  & COME & AEA  & EATA-C$'$ & TCA  & DIPTTA(Ours) \\
        \midrule
        \textbf{ResNet-50(BN)} & 62.0        & 57.7 & 55.1 & 58.1  & 57.6 & 57.1 & 57.2 & 57.3 & 54.6 & 54.2 & 52.9   & 53.4 & \textbf{52.7}  \\
        \textbf{ViT(LN)}       & 47.5        & 45.8 & 41.8 & 42.5  & 43.6 & 45.8 & 39.9 & 45.0 & 38.5 & 37.3 & 35.8   & 36.2 & \textbf{35.3}  \\
        \bottomrule
    \end{tabular}}
    \vspace{-10pt}
\end{table*}

\begin{table*}[t!]
\centering
\caption{An ablation study of the DFPR module and the SCUE module across four datasets, where (C) indicates the CTTA, (BN) indicates employing the ResNet50 as backbone, and (LN) indicates employing ViT as backbone.}
\vspace{-10pt}
    \begin{adjustbox}{width=0.9999\textwidth}
    \resizebox{\textwidth}{!}{
    \begin{tabular}{ccccccccccc}
        \toprule
        \textbf{Prototype} & \textbf{Uncertainty} & \textbf{Example Method} & \textbf{\makecell{CIFAR\\10-C}} & \textbf{\makecell{CIFAR\\100-C}} & \textbf{\makecell{Tiny\\ImageNet-C}} & \textbf{\makecell{ImageNet\\-R(BN)}} & \textbf{\makecell{ImageNet\\-R(LN)}} & \textbf{\makecell{CIFAR\\10-C(C)}} & \textbf{\makecell{CIFAR\\100-C(C)}} & \textbf{\makecell{ImageNet\\-C(C)}} \\
        \midrule

        None & Entropy-based    & Tent  & 18.3      & 31.2       & 63.3           & 57.7           & 43.8           & 20.7           & 60.9            & 62.6           \\
        None & Entropy-based    & COME  & 18.3 & 31.1 & 63.1 & 54.6 & 38.5 & 16.7 & 29.0 & 58.9          \\
        Static     & Entropy-based  & RMT(baseline)     & 18.2      & 31.0       & 62.8           & 57.1           & 45.8           & 16.7           & 30.4            & 60.2           \\
        
        Static     & Source-Calibrated & Ours(w/o DFPR)  & 17.2      & 30.8       & 61.7           & 54.0           & 37.9           & 15.2           & 28.1            & 59.0           \\
        Dynamic(DFPR)     & Entropy-based  & Ours(w/o SCUE) & 16.9      & 30.9       & 61.3           & 53.4           & 38.7           & 14.5           & 28.7            & 59.6            \\
        Dynamic(DFPR) & Source-Calibrated & Ours & \textbf{16.4} & \textbf{30.4} & \textbf{60.1} & \textbf{52.7}           & \textbf{35.3}           & \textbf{13.9} & \textbf{27.8}  & \textbf{58.4} \\
        \bottomrule
    \end{tabular}}
    \end{adjustbox}
    \label{tab:ablation}
    \vspace{-10pt}
\end{table*}

\subsubsection{Comparison on CIFAR100 to CIFAR100-C}
For the more challenging CIFAR100-C dataset, the Source-only model starts with a 46.4\% error rate. It clearly exposes the critical issue of error accumulation in continual test-time adaptation. For example, the Tent-based method suffers from this problem, resulting in a substantially higher error rate of 60.9\% in the long run. Methods like RMT, which employs a symmetric cross-entropy optimization, DSS, which notes the existence of high- and low-quality samples, and BeCoTTA, which improves computational efficiency, all offer incremental progress. On the other hand, the DIPTTA method once again demonstrates dominant performance, reducing the average error to just 27.8\%. This result is a significant improvement over all prior art and underscores its effectiveness in preventing model degradation over long, sequential adaptations.

\subsubsection{Comparison on ImageNet to ImageNet-C}
On the large-scale ImageNet-C dataset, the source-only model struggles immensely, showing an initial error rate of 82.0\%. This benchmark tests the scalability and robustness of adaptation methods under severe corruptions. Although advanced approaches such as CoTTA (62.7\%) provide substantial gains, they still fall short of the top performance. The DIPTTA method again proves its robustness, achieving a leading mean error rate of 58.4\%. It outperforms the latest methods like TCA (59.3\%) and BeCoTTA (60.9\%), confirming its effectiveness on complex, large-scale data.

\subsubsection{Comparison on ImageNet to CCC}
To further assess the robustness of the DIPTTA model in long-sequence continual adaptation tasks, an evaluation using the ImageNet to CCC benchmark was conducted. As detailed in Table~\ref{tab:ccc}, the DIPTTA method shows highly competitive results. Specifically, DIPTTA achieves leading performance on both CCC-Easy and CCC-Medium difficulty levels, with accuracies of 51.2\% and 40.7\%, respectively. Although TCA performs the best on the CCC-Hard setting, DIPTTA obtains the best overall performance when considering the three difficulties, registering an average accuracy of 33.8\%.



\subsection{Comparison with SOTA under Natural Domain Shifts}
As shown in Table~\ref{tab:imagenet-r}, experimental evaluation on the ImageNet-R dataset substantiates the superior and consistent performance of the proposed DIPTTA method. On the ResNet-50 architecture, DIPTTA achieves a state-of-the-art error rate of 52.7\%, and its advantage is even more pronounced on the Vision Transformer (ViT) architecture, where it reduces the error rate to 35.3\%. This consistent top-tier performance validates the potent effectiveness and generalizability of DIPTTA. Its success is attributed to the distilling image prototype, which captures essential and abstract class knowledge rather than just superficial features. This allows the model to effectively distinguish between the core content of an image and its changing artistic style, maintaining strong recognition capabilities while adapting. In contrast, other methods designed primarily for specific algorithmic corruptions struggle to generalize to such complex and varied natural domain shifts.

\begin{table*}[t!]
    \centering
    \caption{Time cost(s) per batch on five datasets. We use a batch size of 200 for all results. We use $'$ to denote the re-implemented.}
     \vspace{-10pt}
    \label{tab:timecost}
    \resizebox{\textwidth}{!}{
    \begin{tabular}{lcccccccccccc}
        \toprule
        \textbf{Dataset} & Source Only & Tent & EATA & MEMO & CoTTA & RMT  & SAR  & COME & AEA  & EATA-C$'$ & TCA  & DIPTTA(Ours) \\
        \midrule
        \textbf{CIFAR10-C} & 0.718        & 1.809 & 2.643 & 62.350  & 3.893  & 2.635 & 3.192 & 2.930 & 31.269 & 2.973   & 6.186 & 2.749  \\
        \textbf{CIFAR100-C}       & 0.827        & 1.992 & 2.862 & 73.403  & 4.127  & 2.965 & 3.428 & 3.357 & 36.707 & 3.505   & 6.539 & 2.979  \\
        \textbf{ImageNet-C}       & 3.157        & 7.093 & 9.561 & 129.139  & 17.285  & 11.773 & 15.021 & 16.153 & 116.250 & 12.379   & 26.154 & 12.235  \\
        \textbf{ImageNet-R(BN)}       & 3.553        & 7.651 & 9.664 & 126.38  & 18.015  & 12.036 & 15.326 & 16.277 & 123.135 & 13.022   & 27.003 & 13.173  \\
        \textbf{ImageNet-R(LN)}       & 10.052        & 22.391 & 28.095 & 376.748  & 73.571  & 37.268 & 47.350 & 50.189 & 392.267 & 43.174   & 98.150 & 39.258  \\
        \bottomrule
    \end{tabular}}
    \vspace{-15pt}
\end{table*}

\subsection{Ablation Studies}

To evaluate the contribution of the two key innovations, a detailed ablation study was performed. Their effectiveness is analyzed by comparing the results with existing methods and by evaluating the standalone impact of each component.

\subsubsection{Effectiveness of Dynamic Feature Prototype Generation}
The results from the CTTA Setting highlight the critical need for dynamic prototypes to prevent catastrophic forgetting. For instance, the Tent method, which lacks a knowledge replay mechanism, sees its error rate on CIFAR100-C increase dramatically from 31.2\% in the TTA setting to 60.9\% under CTTA. More importantly, the DIPTTA method, which uses dynamic prototypes, significantly outperforms RMT, a method that relies on static prototypes, across all datasets. On CIFAR10-C, DIPTTA achieves a 13.9\% error rate compared to an error rate of 16.7\% for RMT. This confirms that dynamically regenerating prototypes that stay aligned with the evolving model is crucial for preventing the knowledge anchor from becoming obsolete.

\subsubsection{Advantage of Source-Calibrated Uncertainty}
The source-calibrated uncertainty mechanism demonstrates clear advantages over conventional entropy-based methods. DIPTTA consistently outperforms methods like Tent and EATA, which are susceptible to overconfident predictions on distribution-shifted data. On the challenging ImageNet-R benchmark, which features natural domain shifts, DIPTTA achieves state-of-the-art results. Furthermore, it accomplishes this while maintaining high computational efficiency, offering a superior trade-off compared to other methods. This performance validates that calibrating uncertainty against stable source knowledge leads to more reliable and robust adaptation.

\subsubsection{Component-wise Analysis}
A direct component-wise ablation, detailed in Table~\ref{tab:ablation}, confirms the individual and combined value of the DFPR and SCUE modules. The baseline model (equivalent to Tent), without either component, records a reference error rate of 18.3\% on CIFAR10-C. Integrating only the {SCUE module} (DIPTTA w/o DFPR) lowers the error rate to 17.2\%, confirming its standalone value in mitigating error accumulation. Similarly, incorporating only the {DFPR module} (DIPTTA w/o SCUE) provides an even greater boost by reducing the error rate to 16.9\%. Most importantly, the {combination of both DFPR and SCUE} in the full DIPTTA model achieves the best result, dropping the CIFAR10-C error rate to 16.4\%, which represents a total improvement of 1.9 percentage points over the baseline. This trend remains consistent across other datasets, such as TinyImageNet-C, where the error rate falls from a baseline of 63.3\% to 60.1\% when both modules are used. The superior performance of the combined system strongly suggests that the two components are complementary and produce a synergistic effect, justifying the inclusion of both in the final model architecture.

\subsection{Further Empirical Analysis}

\subsubsection{Time Cost Analysis}
The time cost analysis in Table~\ref{tab:timecost} confirms the exceptional computational efficiency of DIPTTA. In sharp contrast to computationally intensive methods like MEMO and AEA, whose memory banks and meta-learning render them impractical for time-sensitive applications, DIPTTA maintains a lean profile comparable to its efficient baseline, RMT. The minimal overhead from the proposed DFPR and SCUE modules is the key takeaway; on ImageNet-C, DIPTTA (12.235s) is only marginally slower than RMT (11.773s). This result validates that DIPTTA successfully achieves a superior trade-off, advancing adaptation performance without the cost of inefficiency.

\begin{figure}[!h]
    \centering
    \includegraphics[width=0.9999\linewidth]{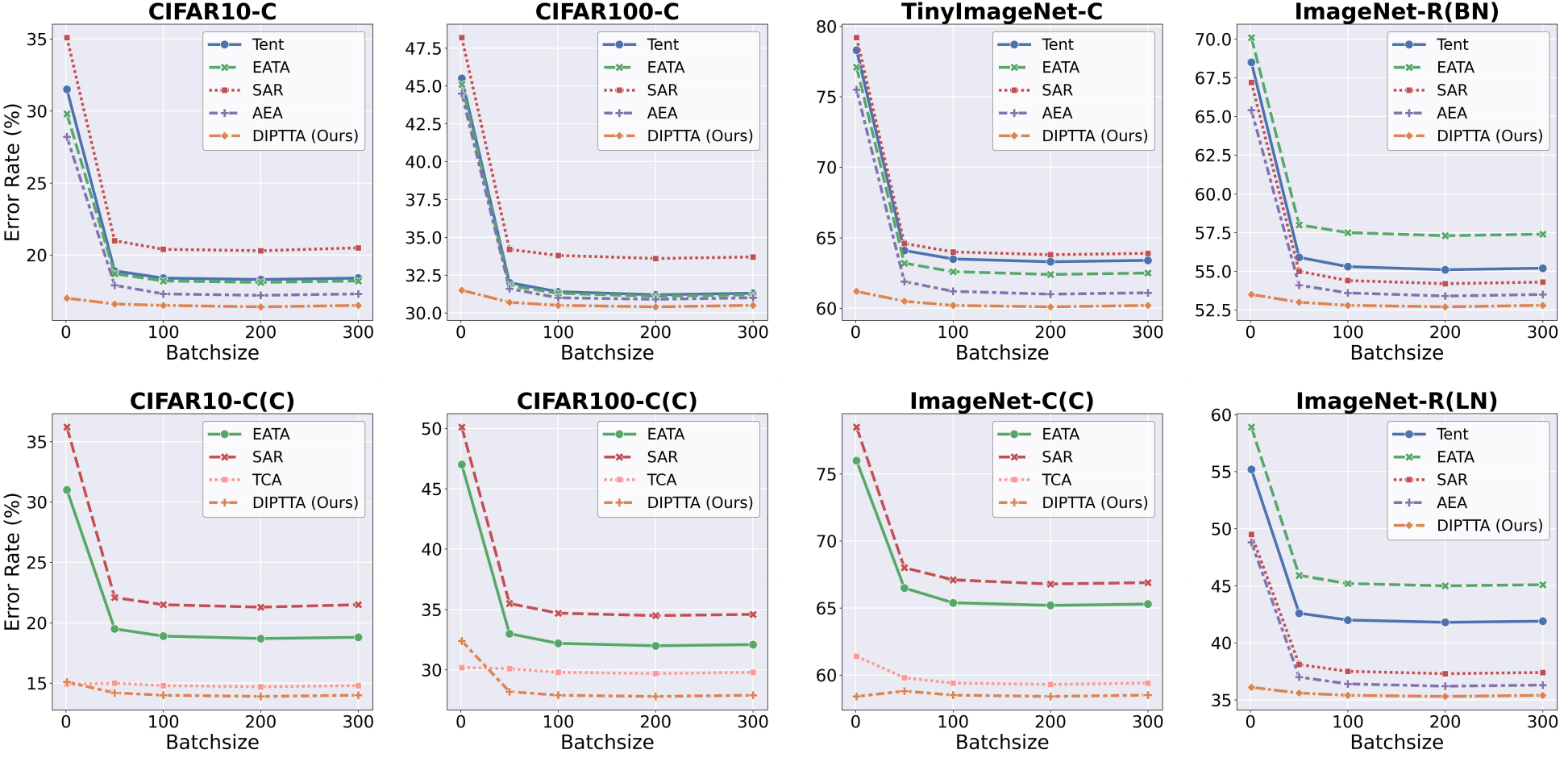}
    \vspace{-15pt}
    \caption{Sensitivity analysis of batch size, where C indicates the CTTA.} 
    \label{fig:bs}
    \vspace{-10pt}
\end{figure}

\subsubsection{Sensitivity Analysis of Batch Size}
As illustrated in Figure~\ref{fig:bs}, DIPTTA exhibits robustness to batch size, consistently outperforming competitors even at a batch size of 1. On CIFAR benchmarks, it achieves optimal error rates (13.9\% and 27.8\%) at batch size 200 while maintaining stability across the 50–300 range, contrasting with the volatility of methods like TCA. This stability extends to ImageNet-C, confirming the low sensitivity and practical reliability of DIPTTA.

\begin{figure}[!h]
    \centering
    \includegraphics[width=0.9999\linewidth]{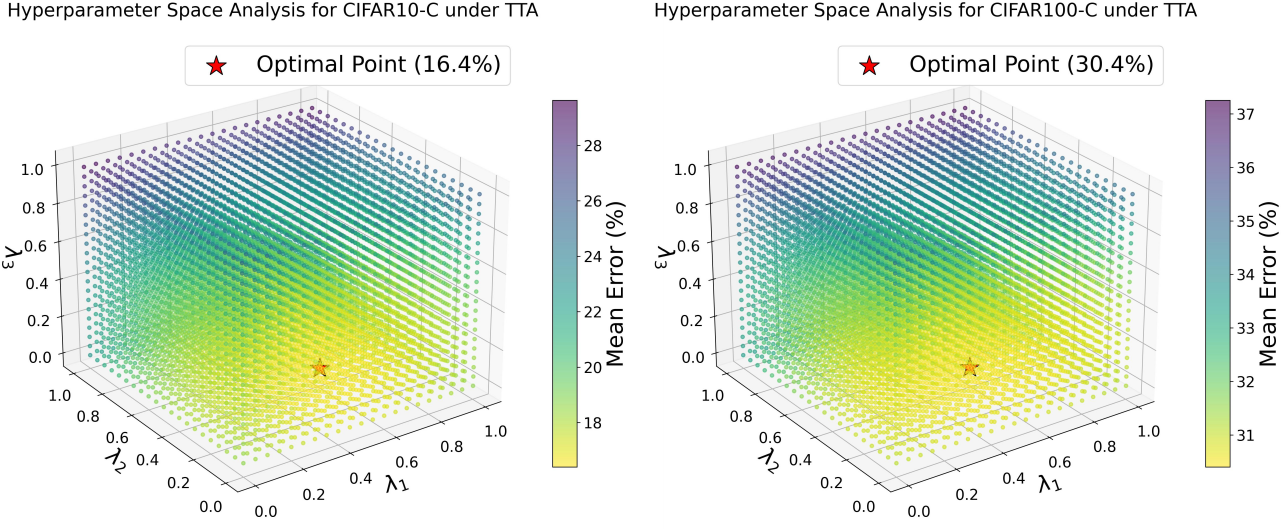}
    \vspace{-15pt}
    \caption{Loss function hyperparameter sensitivity analysis of the TTA method on three benchmark datasets.} 
    \label{fig:hyper}
\end{figure}

\subsubsection{Hyperparameter Sensitivity Analysis}
\label{sec:Hyperparameter}
Figure~\ref{fig:hyper} illustrates the impact of three hyperparameters ($\lambda_1\lambda_2,\lambda_3$) on the performance of a TTA method across the CIFAR10-C and CIFAR100-C datasets. The color of each point represents the mean error rate (\%), where brighter yellow indicates lower error. The optimal points for the datasets were found to be (0.5, 0.25, 0.15) and (0.45, 0.2, 0.2), respectively. Given the proximity of these optima, the single parameter set of (0.5, 0.25, 0.15), marked by the red star, is selected for all experiments.

\begin{figure}[!t]
 \centering
 \includegraphics[width=0.99\linewidth, trim={0cm 0 0cm 0}, clip]{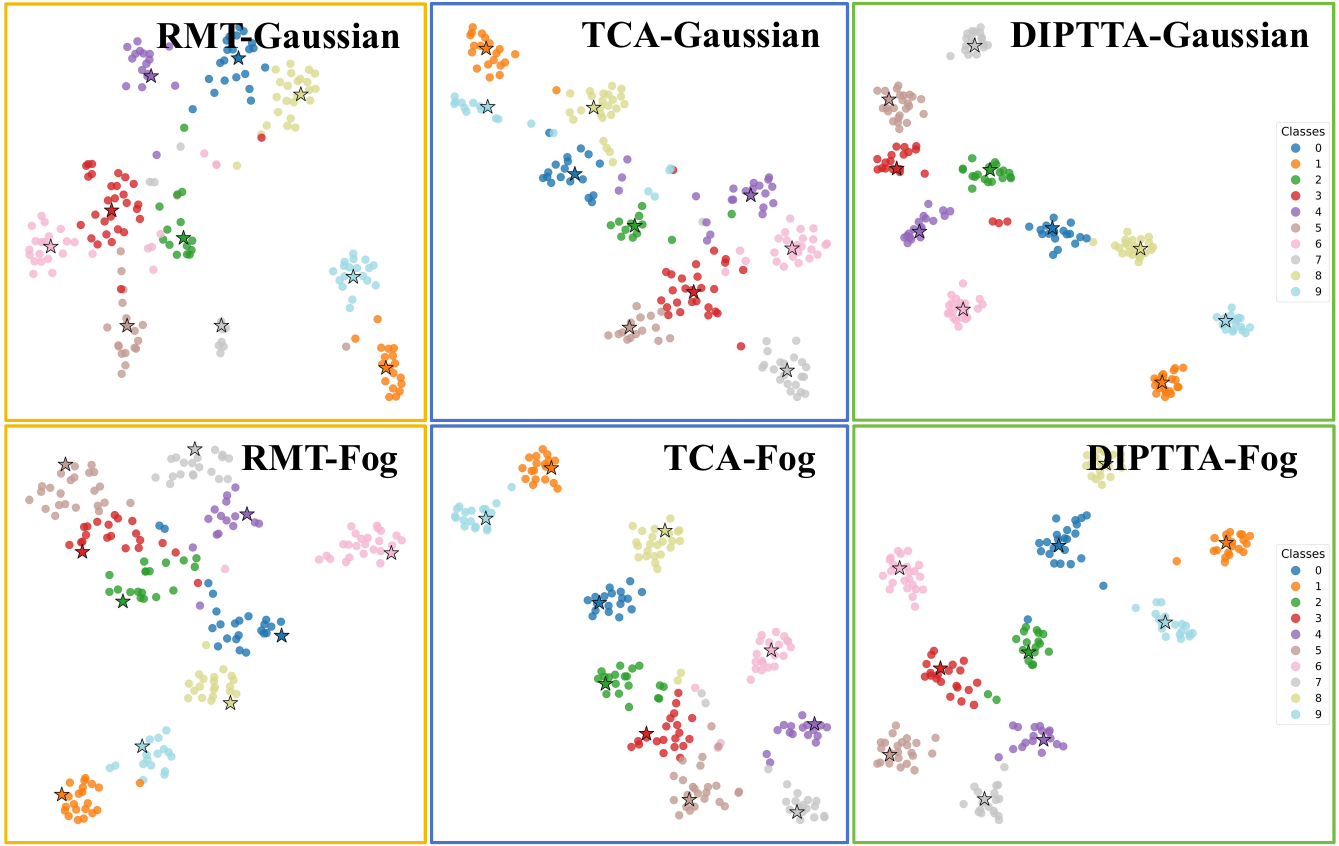} 
 \vspace{-10pt}
 \caption{Visualization of t-SNE for two methods and our proposed DIPTTA on CIFAR10-C(TTA Scenarios). We select a random batch of features from the Gaussian and Fog domains for comparative visualization.}
\label{fig.tsne}
\end{figure}

\subsubsection{t-SNE Analysis}
t-SNE~\cite{van2008visualizing} is utilized to visualize the feature representations of the model. As shown in Fig.~\ref{fig.tsne}, the feature vectors (dots) from the DIPTTA model exhibit significantly tighter clustering around the source domain prototypes (stars) compared to other methods on CIFAR10-C. This improved performance stems from two key factors: 1) the employed data distillation technique refines source domain knowledge to create more accurate class prototypes that better represent true feature centroids; and 2) the combined uncertainty-weighted consistency and contrastive losses form a robust constraint that aligns target feature vectors with these optimized prototypes. Consequently, the model learns more compact and separable features, creating clearer decision boundaries.

\section{CONCLUSION}

In this paper, the critical challenges of error accumulation from noisy pseudo-labels and catastrophic forgetting of source knowledge during test-time adaptation are analyzed. To this end, the DIPTTA framework is proposed, which introduces a Distilling Image Prototype (DIP) as a compact and regenerative source knowledge anchor. The core contribution is twofold. First, the DIP enables dynamic feature prototype replay, which continuously regenerates feature prototypes aligned with the evolving feature space of the model, effectively overcoming the misalignment issue of static prototypes. Second, it facilitates source-calibrated uncertainty estimation, which leverages the stable DIP to debias uncertainty quantification and reliably suppress error propagation. Extensive experiments on multiple benchmarks validate that DIPTTA significantly outperforms state-of-the-art methods. The results demonstrate that the proposed approach provides a robust and effective solution for model adaptation in non-stationary test environments.

\bibliographystyle{IEEEtran}
\bibliography{main}{}

@String(ICLR = {Int. Conf. Learn. Represent.})

@String(IJCAI = {IJCAI})

@String(ICLR  = {ICLR})

@inproceedings{wang2020tent,
  title={Tent: Fully Test-Time Adaptation by Entropy Minimization},
  author={Wang, Dequan and Shelhamer, Evan and Liu, Shaoteng and Olshausen, Bruno and Darrell, Trevor},
  booktitle={International Conference on Learning Representations},
  year={2020}
}

@inproceedings{ma2024improved,
  title={Improved self-training for test-time adaptation},
  author={Ma, Jing},
  booktitle={Proceedings of the IEEE/CVF Conference on Computer Vision and Pattern Recognition},
  pages={23701--23710},
  year={2024}
}

@inproceedings{niu2023towards,
  title={Towards Stable Test-time Adaptation in Dynamic Wild World},
  author={Niu, Shuaicheng and Wu, Jiaxiang and Zhang, Yifan and Wen, Zhiquan and Chen, Yaofo and Zhao, Peilin and Tan, Mingkui},
  booktitle={The Eleventh International Conference on Learning Representations},
  year={2023}
}

@article{liang2025comprehensive,
  title={A comprehensive survey on test-time adaptation under distribution shifts},
  author={Liang, Jian and He, Ran and Tan, Tieniu},
  journal={International Journal of Computer Vision},
  volume={133},
  number={1},
  pages={31--64},
  year={2025},
  publisher={Springer}
}

@article{goyal2022test,
  title={Test time adaptation via conjugate pseudo-labels},
  author={Goyal, Sachin and Sun, Mingjie and Raghunathan, Aditi and Kolter, J Zico},
  journal={Advances in Neural Information Processing Systems},
  volume={35},
  pages={6204--6218},
  year={2022}
}

@inproceedings{sinha2023test,
  title={Test: Test-time self-training under distribution shift},
  author={Sinha, Samarth and Gehler, Peter and Locatello, Francesco and Schiele, Bernt},
  booktitle={Proceedings of the IEEE/CVF Winter Conference on Applications of Computer Vision},
  pages={2759--2769},
  year={2023}
}

@inproceedings{jang2022test,
  title={Test-Time Adaptation via Self-Training with Nearest Neighbor Information},
  author={Jang, Minguk and Chung, Sae-Young and Chung, Hye Won},
  booktitle={The Eleventh International Conference on Learning Representations},
  year={2022}
}

@article{dong2025certaintta,
  title={CertainTTA: Estimating uncertainty for test-time adaptation on medical image segmentation},
  author={Dong, Xingbo and Wang, Liwen and Lv, Xingguo and Zhang, Xiaoyan and Zhang, Hui and Pu, Bin and Gao, Zhan and Liao, Iman Yi and Jin, Zhe},
  journal={Information Fusion},
  pages={103300},
  year={2025},
  publisher={Elsevier}
}

@article{tan2025uncertainty,
  title={Uncertainty-calibrated test-time model adaptation without forgetting},
  author={Tan, Mingkui and Chen, Guohao and Wu, Jiaxiang and Zhang, Yifan and Chen, Yaofo and Zhao, Peilin and Niu, Shuaicheng},
  journal={IEEE Transactions on Pattern Analysis and Machine Intelligence},
  year={2025},
  publisher={IEEE}
}

@article{tarvainen2017mean,
  title={Mean teachers are better role models: Weight-averaged consistency targets improve semi-supervised deep learning results},
  author={Tarvainen, Antti and Valpola, Harri},
  journal={Advances in neural information processing systems},
  volume={30},
  year={2017}
}

@inproceedings{dobler2023robust,
  title={Robust mean teacher for continual and gradual test-time adaptation},
  author={D{\"o}bler, Mario and Marsden, Robert A and Yang, Bin},
  booktitle={Proceedings of the IEEE/CVF Conference on Computer Vision and Pattern Recognition},
  pages={7704--7714},
  year={2023}
}

@inproceedings{chen2022contrastive,
  title={Contrastive test-time adaptation},
  author={Chen, Dian and Wang, Dequan and Darrell, Trevor and Ebrahimi, Sayna},
  booktitle={Proceedings of the IEEE/CVF Conference on Computer Vision and Pattern Recognition},
  pages={295--305},
  year={2022}
}

@inproceedings{wang2022continual,
  title={Continual test-time domain adaptation},
  author={Wang, Qin and Fink, Olga and Van Gool, Luc and Dai, Dengxin},
  booktitle={Proceedings of the IEEE/CVF Conference on Computer Vision and Pattern Recognition},
  pages={7201--7211},
  year={2022}
}

@inproceedings{wang2024continual,
  title={Continual test-time domain adaptation via dynamic sample selection},
  author={Wang, Yanshuo and Hong, Jie and Cheraghian, Ali and Rahman, Shafin and Ahmedt-Aristizabal, David and Petersson, Lars and Harandi, Mehrtash},
  booktitle={Proceedings of the IEEE/CVF Winter Conference on Applications of Computer Vision},
  pages={1701--1710},
  year={2024}
}

@inproceedings{lee2024becotta,
  title={BECoTTA: Input-dependent Online Blending of Experts for Continual Test-time Adaptation},
  author={Lee, Daeun and Yoon, Jaehong and Hwang, Sung Ju},
  booktitle={International Conference on Machine Learning},
  pages={27072--27093},
  year={2024},
  organization={PMLR}
}

@article{krizhevsky2009learning,
  title={Learning multiple layers of features from tiny images},
  author={Krizhevsky, Alex and Hinton, Geoffrey and others},
  year={2009},
  publisher={Toronto, ON, Canada}
}

@inproceedings{zagoruyko2016wide,
  title={Wide Residual Networks},
  author={Zagoruyko, Sergey and Komodakis, Nikos},
  booktitle={British Machine Vision Conference 2016},
  year={2016},
  organization={British Machine Vision Association}
}

@article{yin2019fourier,
  title={A fourier perspective on model robustness in computer vision},
  author={Yin, Dong and Gontijo Lopes, Raphael and Shlens, Jon and Cubuk, Ekin Dogus and Gilmer, Justin},
  journal={Advances in Neural Information Processing Systems},
  volume={32},
  year={2019}
}

@inproceedings{croce2020robustbench,
  title={RobustBench: a standardized adversarial robustness benchmark},
  author={Croce, Francesco and Andriushchenko, Maksym and Sehwag, Vikash and Debenedetti, Edoardo and Flammarion, Nicolas and Chiang, Mung and Mittal, Prateek and Hein, Matthias},
  booktitle={Thirty-fifth Conference on Neural Information Processing Systems Datasets and Benchmarks Track (Round 2)},
  year={2020}
}

@inproceedings{carta2022distilled,
  title={Distilled Replay: Overcoming Forgetting Through Synthetic Samples},
  author={Carta, Antonio and Cossu, Andrea and Lomonaco, Vincenzo and Bacciu, Davide},
  booktitle={Continual Semi-Supervised Learning: First International Workshop, CSSL 2021, Virtual Event, August 19-20, 2021, Revised Selected Papers},
  volume={13418},
  pages={104},
  year={2022},
  organization={Springer Nature}
}

@article{chakrabarty2023santa,
  title={Santa: Source anchoring network and target alignment for continual test time adaptation},
  author={Chakrabarty, Goirik and Sreenivas, Manogna and Biswas, Soma},
  journal={Transactions on Machine Learning Research},
  year={2023}
}

@inproceedings{niu2022efficient,
  title={Efficient test-time model adaptation without forgetting},
  author={Niu, Shuaicheng and Wu, Jiaxiang and Zhang, Yifan and Chen, Yaofo and Zheng, Shijian and Zhao, Peilin and Tan, Mingkui},
  booktitle={International conference on machine learning},
  pages={16888--16905},
  year={2022},
  organization={PMLR}
}

@article{press2023rdumb,
  title={Rdumb: A simple approach that questions our progress in continual test-time adaptation},
  author={Press, Ori and Schneider, Steffen and K{\"u}mmerer, Matthias and Bethge, Matthias},
  journal={Advances in Neural Information Processing Systems},
  volume={36},
  pages={39915--39935},
  year={2023}
}

@article{van2008visualizing,
  title={Visualizing data using t-SNE.},
  author={Van der Maaten, Laurens and Hinton, Geoffrey},
  journal={Journal of machine learning research},
  volume={9},
  number={11},
  year={2008}
}

@article{denker1990transforming,
  title={Transforming neural-net output levels to probability distributions},
  author={Denker, John and LeCun, Yann},
  journal={Advances in neural information processing systems},
  volume={3},
  year={1990}
}

@book{mackay1992bayesian,
  title={Bayesian methods for adaptive models},
  author={Mackay, David John Cameron},
  year={1992},
  publisher={California Institute of Technology}
}

@article{rahaman2021uncertainty,
  title={Uncertainty quantification and deep ensembles},
  author={Rahaman, Rahul and others},
  journal={Advances in neural information processing systems},
  volume={34},
  pages={20063--20075},
  year={2021}
}

@article{abe2022deep,
  title={Deep ensembles work, but are they necessary?},
  author={Abe, Taiga and Buchanan, Estefany Kelly and Pleiss, Geoff and Zemel, Richard and Cunningham, John P},
  journal={Advances in Neural Information Processing Systems},
  volume={35},
  pages={33646--33660},
  year={2022}
}

@inproceedings{choiadaptive,
  title={Adaptive Energy Alignment for Accelerating Test-Time Adaptation},
  author={Choi, Wonjeong and Kim, Do-Yeon and Park, Jungwuk and Lee, Jungmoon and Park, Younghyun and Han, Dong-Jun and Moon, Jaekyun},
  booktitle={The Thirteenth International Conference on Learning Representations},
  year={2025}
}

@inproceedings{liang2020we,
  title={Do we really need to access the source data? source hypothesis transfer for unsupervised domain adaptation},
  author={Liang, Jian and Hu, Dapeng and Feng, Jiashi},
  booktitle={Proceedings of the International Conference on Machine Learning (ICML)},
  pages={6028--6039},
  year={2020},
}

@inproceedings{gal2016dropout,
  title={Dropout as a {B}ayesian approximation: {R}epresenting model uncertainty in deep learning},
  author={Gal, Yarin and Ghahramani, Zoubin},
  booktitle={Proceedings of the International Conference on Machine Learning (ICML)},
  pages={1050--1059},
  year={2016},
}

@article{xiao2024beyond,
  title={Beyond model adaptation at test time: A survey},
  author={Xiao, Zehao and Snoek, Cees GM},
  journal={arXiv preprint arXiv:2411.03687},
  year={2024}
}

@article{wang2025search,
  title={In search of lost online test-time adaptation: A survey},
  author={Wang, Zixin and Luo, Yadan and Zheng, Liang and Chen, Zhuoxiao and Wang, Sen and Huang, Zi},
  journal={International Journal of Computer Vision},
  volume={133},
  number={3},
  pages={1106--1139},
  year={2025},
  publisher={Springer}
}

@inproceedings{jin2025fedwsidd,
  title={Fedwsidd: Federated whole slide image classification via dataset distillation},
  author={Jin, Haolong and Liu, Shenglin and Cong, Cong and Feng, Qingmin and Liu, Yongzhi and Huang, Lina and Hu, Yingzi},
  booktitle={International Conference on Medical Image Computing and Computer-Assisted Intervention},
  pages={178--188},
  year={2025},
  organization={Springer}
}

@inproceedings{geng2023survey,
  title={A Survey on Dataset Distillation: Approaches, Applications and Future Directions},
  author={Geng, Jiahui and Chen, Zongxiong and Wang, Yuandou and Woisetschlaeger, Herbert and Schimmler, Sonja and Mayer, Ruben and Zhao, Zhiming and Rong, Chunming},
  booktitle={IJCAI},
  year={2023}
}

@article{zhang2022memo,
  title={Memo: Test time robustness via adaptation and augmentation},
  author={Zhang, Marvin and Levine, Sergey and Finn, Chelsea},
  journal={Advances in neural information processing systems},
  volume={35},
  pages={38629--38642},
  year={2022}
}

@inproceedings{zhang2025come,
  title = {COME: Test-Time Adaptation by Conservatively Minimizing Entropy},
  author = {Qingyang Zhang and Yatao Bian and Xinke Kong and Peilin Zhao and Changqing Zhang},
  booktitle = {International Conference on Learning Representations},
  year = {2025}
}

@inproceedings{ritter2018scalable,
  title={A scalable laplace approximation for neural networks},
  author={Ritter, Hippolyt and Botev, Aleksandar and Barber, David},
  booktitle={6th international conference on learning representations, ICLR 2018-conference track proceedings},
  volume={6},
  year={2018},
  organization={International Conference on Representation Learning}
}

@inproceedings{hendrycks2021many,
  title={The many faces of robustness: A critical analysis of out-of-distribution generalization},
  author={Hendrycks, Dan and Basart, Steven and Mu, Norman and Kadavath, Saurav and Wang, Frank and Dorundo, Evan and Desai, Rahul and Zhu, Tyler and Parajuli, Samyak and Guo, Mike and others},
  booktitle={Proceedings of the IEEE/CVF international conference on computer vision},
  pages={8340--8349},
  year={2021}
}

@inproceedings{kristiadi2020being,
  title={Being bayesian, even just a bit, fixes overconfidence in relu networks},
  author={Kristiadi, Agustinus and Hein, Matthias and Hennig, Philipp},
  booktitle={International conference on machine learning},
  pages={5436--5446},
  year={2020},
  organization={PMLR}
}

@article{daxberger2021laplace,
  title={Laplace redux-effortless bayesian deep learning},
  author={Daxberger, Erik and Kristiadi, Agustinus and Immer, Alexander and Eschenhagen, Runa and Bauer, Matthias and Hennig, Philipp},
  journal={Advances in neural information processing systems},
  volume={34},
  pages={20089--20103},
  year={2021}
}

@article{iwasawa2021test,
  title={Test-time classifier adjustment module for model-agnostic domain generalization},
  author={Iwasawa, Yusuke and Matsuo, Yutaka},
  journal={Advances in Neural Information Processing Systems},
  volume={34},
  pages={2427--2440},
  year={2021}
}

@inproceedings{hendrycks2016baseline,
  title={A Baseline for Detecting Misclassified and Out-of-Distribution Examples in Neural Networks},
  author={Hendrycks, Dan and Gimpel, Kevin},
  booktitle={International Conference on Learning Representations},
  year={2017}
}

@article{liu2020energy,
  title={Energy-based out-of-distribution detection},
  author={Liu, Weitang and Wang, Xiaoyun and Owens, John and Li, Yixuan},
  journal={Advances in neural information processing systems},
  volume={33},
  pages={21464--21475},
  year={2020}
}

@inproceedings{bottou2010large,
  title={Large-scale machine learning with stochastic gradient descent},
  author={Bottou, L{\'e}on},
  booktitle={Proceedings of COMPSTAT},
  pages={177--186},
  year={2010},
  organization={Springer}
}

@inproceedings{zhao2021dataset,
  title={Dataset condensation with differentiable siamese augmentation},
  author={Zhao, Bo and Bilen, Hakan},
  booktitle={International Conference on Machine Learning},
  pages={12674--12685},
  year={2021},
  organization={PMLR}
}

@article{trimmer2011decision,
  title={Decision-making under uncertainty: biases and Bayesians},
  author={Trimmer, Pete C and Houston, Alasdair I and Marshall, James AR and Mendl, Mike T and Paul, Elizabeth S and McNamara, John M},
  journal={Animal cognition},
  volume={14},
  number={4},
  pages={465--476},
  year={2011},
  publisher={Springer}
}

@article{kumar2019verified,
  title={Verified uncertainty calibration},
  author={Kumar, Ananya and Liang, Percy S and Ma, Tengyu},
  journal={Advances in neural information processing systems},
  volume={32},
  year={2019}
}

@inproceedings{yuan2023robust,
  title={Robust test-time adaptation in dynamic scenarios},
  author={Yuan, Longhui and Xie, Binhui and Li, Shuang},
  booktitle={Proceedings of the IEEE/CVF Conference on Computer Vision and Pattern Recognition},
  pages={15922--15932},
  year={2023}
}

@inproceedings{boudiaf2022parameter,
  title={Parameter-free online test-time adaptation},
  author={Boudiaf, Malik and Mueller, Romain and Ben Ayed, Ismail and Bertinetto, Luca},
  booktitle={Proceedings of the IEEE/CVF Conference on Computer Vision and Pattern Recognition},
  pages={8344--8353},
  year={2022}
}

@inproceedings{wang2023feature,
  title={Feature alignment and uniformity for test time adaptation},
  author={Wang, Shuai and Zhang, Daoan and Yan, Zipei and Zhang, Jianguo and Li, Rui},
  booktitle={Proceedings of the IEEE/CVF Conference on Computer Vision and Pattern Recognition},
  pages={20050--20060},
  year={2023}
}

@inproceedings{jung2023cafa,
  title={Cafa: Class-aware feature alignment for test-time adaptation},
  author={Jung, Sanghun and Lee, Jungsoo and Kim, Nanhee and Shaban, Amirreza and Boots, Byron and Choo, Jaegul},
  booktitle={Proceedings of the IEEE/CVF International Conference on Computer Vision},
  pages={19060--19071},
  year={2023}
}

@article{zhang2025test,
  title={Test-time adaptation for object detection via Dynamic Dual Teaching},
  author={Zhang, Siqi and Zhang, Lu and Liu, Zhiyong},
  journal={Image and Vision Computing},
  pages={105740},
  year={2025},
  publisher={Elsevier}
}

@inproceedings{sojka2023ar,
  title={Ar-tta: A simple method for real-world continual test-time adaptation},
  author={S{\'o}jka, Damian and Cygert, Sebastian and Twardowski, Bart{\l}omiej and Trzci{\'n}ski, Tomasz},
  booktitle={Proceedings of the IEEE/CVF International Conference on Computer Vision},
  pages={3491--3495},
  year={2023}
}

@article{gong2022note,
  title={Note: Robust continual test-time adaptation against temporal correlation},
  author={Gong, Taesik and Jeong, Jongheon and Kim, Taewon and Kim, Yewon and Shin, Jinwoo and Lee, Sung-Ju},
  journal={Advances in Neural Information Processing Systems},
  volume={35},
  pages={27253--27266},
  year={2022}
}

@inproceedings{niloy2024effective,
  title={Effective restoration of source knowledge in continual test time adaptation},
  author={Niloy, Fahim Faisal and Ahmed, Sk Miraj and Raychaudhuri, Dripta S and Oymak, Samet and Roy-Chowdhury, Amit K},
  booktitle={Proceedings of the IEEE/CVF Winter Conference on Applications of Computer Vision},
  pages={2091--2100},
  year={2024}
}

@article{ye2025domain,
  title={Domain consistency learning for continual test-time adaptation in image semantic segmentation},
  author={Ye, Yanyu and Wei, Wei and Zhang, Lei and Ding, Chen and Zhang, Yanning},
  journal={Pattern Recognition},
  volume={165},
  pages={111585},
  year={2025},
  publisher={Elsevier}
}

@inproceedings{chen2023improved,
  title={Improved test-time adaptation for domain generalization},
  author={Chen, Liang and Zhang, Yong and Song, Yibing and Shan, Ying and Liu, Lingqiao},
  booktitle={Proceedings of the IEEE/CVF Conference on Computer Vision and Pattern Recognition},
  pages={24172--24182},
  year={2023}
}

@article{wang2025decoupled,
  title={Decoupled prototype learning for reliable test-time adaptation},
  author={Wang, Guowei and Ding, Changxing and Tan, Wentao and Tan, Mingkui},
  journal={IEEE Transactions on Multimedia},
  year={2025},
  publisher={IEEE}
}

@article{ma2024discrepancy,
  title={Discrepancy and structure-based contrast for test-time adaptive retrieval},
  author={Ma, Zeyu and Li, Yuqi and Luo, Yizhi and Luo, Xiao and Li, Jinxing and Chen, Chong and Hua, Xian-Sheng and Lu, Guangming},
  journal={IEEE Transactions on Multimedia},
  volume={26},
  pages={8665--8677},
  year={2024},
  publisher={IEEE}
}

@article{tan2025source,
  title={Source-Free Elastic Model Adaptation for Vision-and-Language Navigation},
  author={Tan, Mingkui and Chen, Peihao and Zhi, Hongyan and Mai, Jiajie and Rosman, Benjamin and Ji, Dongyu and Zeng, Runhao},
  journal={IEEE Transactions on Multimedia},
  year={2025},
  publisher={IEEE}
}

@article{hu2023unleashing,
  title={Unleashing knowledge potential of source hypothesis for source-free domain adaptation},
  author={Hu, Bingyu and Liu, Jiawei and Zheng, Kecheng and Zha, Zheng-Jun},
  journal={IEEE Transactions on Multimedia},
  volume={26},
  pages={5422--5434},
  year={2023},
  publisher={IEEE}
}

@article{tian2025high,
  title={High Specificity Guided Cross-Domain Few-Shot Segmentation},
  author={Tian, Pinzhuo and Zhou, Xiaojie and Yu, Hang and Xie, Shaorong},
  journal={IEEE Transactions on Multimedia},
  year={2025},
  publisher={IEEE}
}

\end{document}